\documentclass[journal]{IEEEtran}
\usepackage{times}
\usepackage{color}
\usepackage{graphicx}
\usepackage{amsmath}
\usepackage{amssymb}
\usepackage{tabularx}
\usepackage{bbm}
\usepackage{multirow}
\usepackage{arydshln}
\usepackage{bigstrut}
\usepackage{cite}
\usepackage{pbox}
\usepackage{makecell}
\usepackage{subfigure}

\begin{document}
%
\title{Few-Shot Learning with Geometric Constraints}
%
%
%

\author{Hong-Gyu~Jung and Seong-Whan~Lee,~\IEEEmembership{Fellow,~IEEE \vspace{-0.7\baselineskip}}
\thanks{\copyright 2020 IEEE. Personal use of this material is permitted. Permission
from IEEE must be obtained for all other uses, in any current or future
media, including reprinting/republishing this material for advertising or
promotional purposes, creating new collective works, for resale or
redistribution to servers or lists, or reuse of any copyrighted
component of this work in other works. DOI: 10.1109/TNNLS.2019.2957187}
\thanks{This work was supported by Institute of Information \& communications Technology Planning \& Evaluation (IITP) grant funded by the Korea government(MSIT) (No. 2017-0-01779, A machine learning and statistical inference framework for explainable artificial intelligence, No. 2019-0-01371, Development of brain-inspired AI with human-like intelligence, and No. 2019-0-00079, Fostering high-quality artificial intelligence talent)}
\thanks{H.-G. Jung is with the Department of Brain and Cognitive Engineering and S.-W. Lee is with the Department of Artificial Intelligence, Korea University, Anam-dong, Seongbuk-ku, Seoul 02841, Korea. e-mail: \{hkjung00, sw.lee\}@korea.ac.kr (Corresponding author: Seong-Whan Lee). \vspace{-2.0\baselineskip}}
}

\markboth{IEEE Transactions on Neural Networks and Learning Systems, 2020}%
{Shell \MakeLowercase{\textit{et al.}}: Bare Demo of IEEEtran.cls for IEEE Journals}

\IEEEoverridecommandlockouts
\maketitle

\begin{abstract}
In this paper, we consider the problem of few-shot learning for classification. We assume a network trained for base categories with a large number of training examples, and we aim to add novel categories to it that have only a few, e.g., one or five, training examples. This is a challenging scenario because (1) high performance is required in both the base and novel categories, and (2) training the network for the new categories with few training examples can contaminate the feature space trained well for the base categories. To address these challenges, we propose two geometric constraints to fine-tune the network with a few training examples. The first constraint enables features of the novel categories to cluster near the category weights, and the second maintains the weights of the novel categories far from the weights of the base categories. By applying the proposed constraints, we extract discriminative features for the novel categories while preserving the feature space learned for the base categories. Using public datasets for few-shot learning that are subsets of ImageNet, we demonstrate that the proposed method outperforms prevalent methods by a large margin.
\end{abstract}


%
\IEEEpeerreviewmaketitle

\section{Introduction}
\IEEEPARstart{D} {\uppercase{eep}} networks have achieved state-of-the art performance in a variety of fields, such as visual recognition \cite{he2016deep, huang2017densely, szegedy2015going, chen2017deep, cao2018data}, object detection  \cite{ren2015faster} and semantic segmentation \cite{long2015fully}. Compared to previous studies \cite{arbelaez2012semantic, sanchez2013image, swlee_1, swlee_2, swlee_3}, these impressive achievements are primarily owing to the availability of large numbers of training examples. For instance, current deep networks for image classification match human performance but require a large number of training examples \cite{he2016deep}. Humans can quickly and accurately recognize novel categories using only one example based on previous knowledge \cite{schmidt2009meaning}. Mimicking the cognitive abilities of humans is difficult, as the layers of deep networks are heavily optimized for the trained categories. In particular, the top layers tend to learn category-specific features \cite{zeiler2014visualizing}, which are not flexible enough to learn novel categories without forgetting the parameters learned before. Thus, to continually add novel categories, the network must be re-trained by using a large number of training examples for the base and novel categories \cite{Kirkpatrick3521}. Apart from the complexity of training, building large numbers of training examples requires a significant amount of human effort and incurs high costs.

To solve this problem, few-shot learning, which can train neural networks using a few training examples, has attracted considerable research interest. In this paper, we consider a network problem where its base categories have been trained using a large number of training examples and novel categories with a few, e.g., one or five, training examples are to be added. This problem is challenging as the network needs to exhibit high performance in both the base and novel categories. In other words, an effective algorithm needs to be able to generate discriminative features and weights\footnote{In this paper, we alternately denote the output of a feature extractor as features and activations. Weights are referred to as category weights used in a classifier.\vspace{-6mm}} for the novel categories without affecting those of the base categories.

\begin{figure*}
        \subfigure[Trained for base categories]
        {
                \centering
                \includegraphics[width=.25\linewidth]{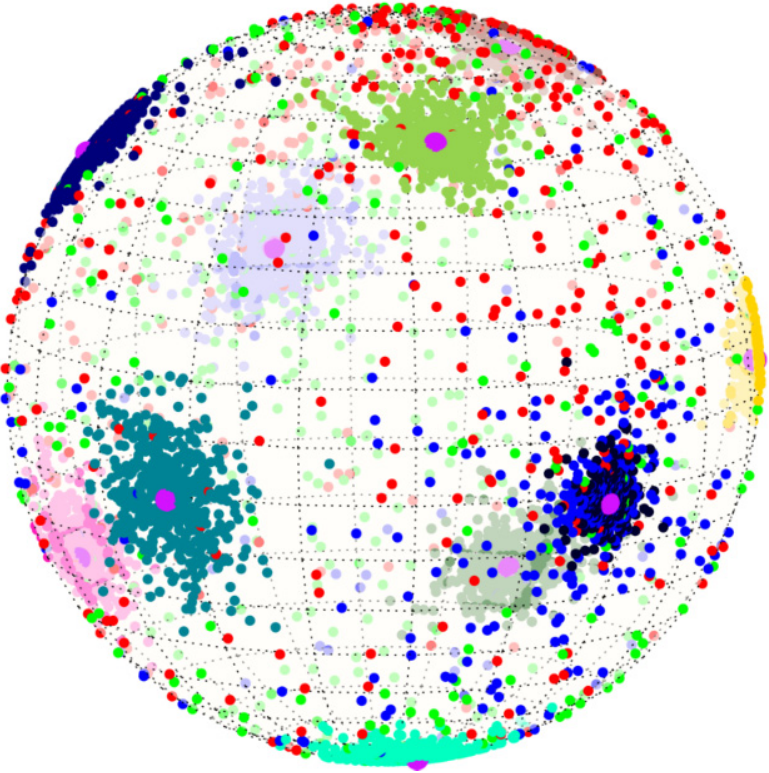}
                \label{fig:trained}
		}
        \subfigure[Fine-tuned by Softmax]
        {
                \centering
                \includegraphics[width=.25\linewidth]{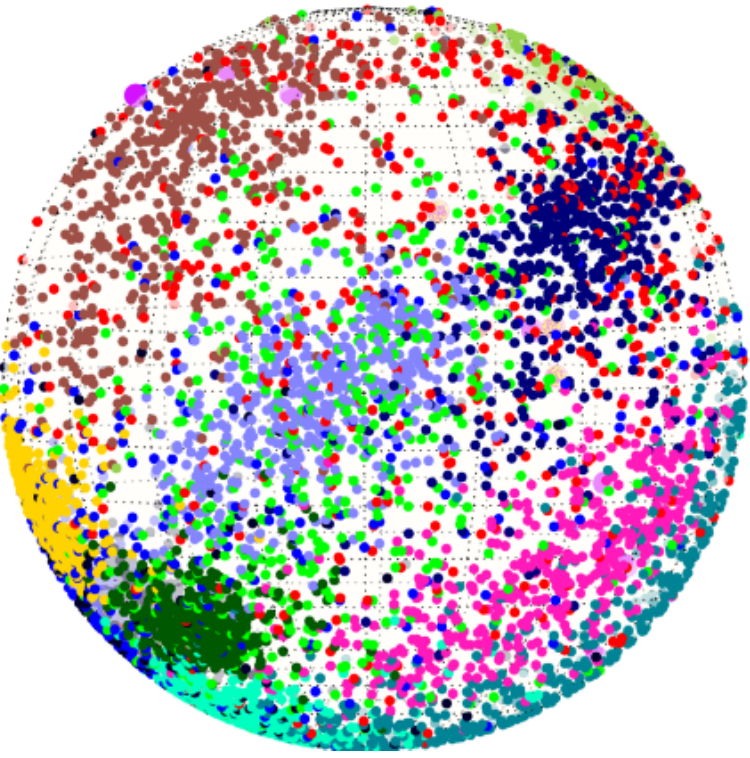}
                \label{fig:finetuend}
		}
        \subfigure[Fine-tuned by geometric constraints]
        {
                \centering
                \includegraphics[width=.25\linewidth]{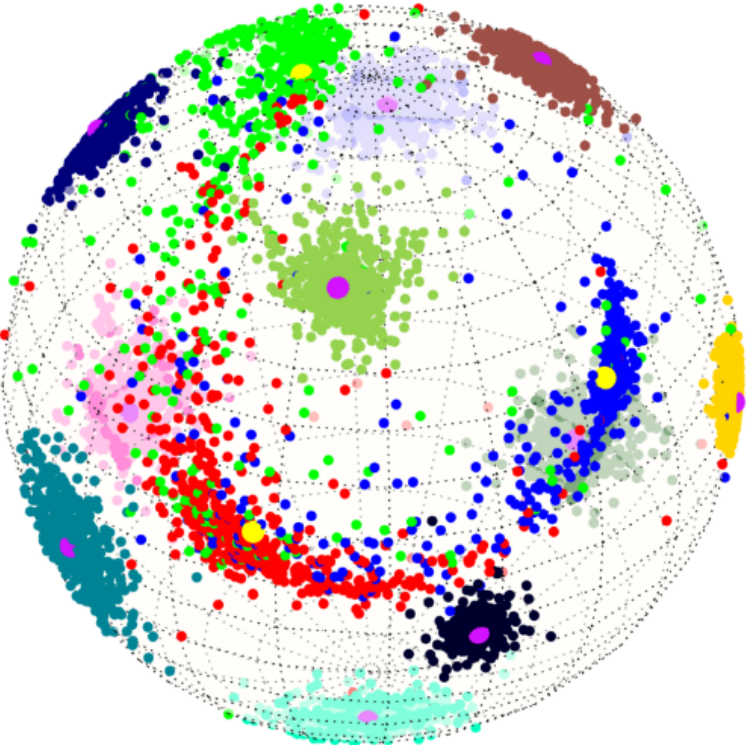}
                \label{fig:proposed}
		}
        \subfigure
        {
                \centering
				\raisebox{10pt}{
                \includegraphics[width=0.10\linewidth]{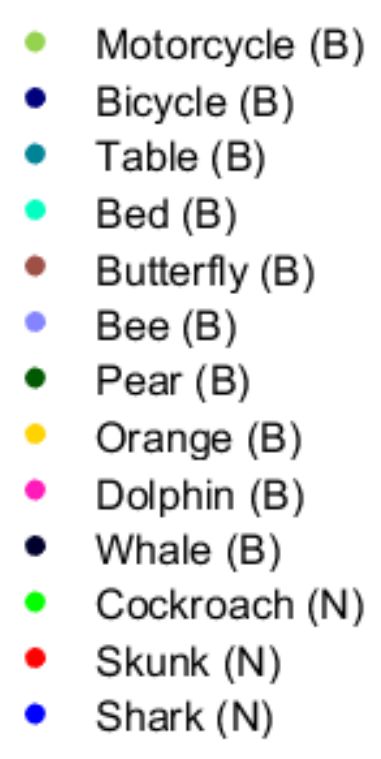}}
                \label{fig:CIFAR100_labels}
		}
        \caption{(Best viewed in color) Examples of base and novel categories in the feature space. We trained 10 base categories chosen from the CIFAR 100 dataset \cite{krizhevsky2009learning} using $500$ training examples for each base category. (a) shows features extracted from both base and novel categories. The base categories are clustered near the weights (hot pink). ``Shark,'' one of the novel categories, is clustered well but its location overlaps with that of ``Whale''. The other two novel categories are spread out. (b) shows the results after the network was fine-tuned with five training examples for each novel category. For the network, we added classification weights for novel categories to the classifier and froze the classification weights for base categories during fine-tuning. We did not use any training examples for base categories. (c) demonstrates that our proposed geometric constraints can locate the features of the novel categories discriminatively and preserve the feature space of the base categories. The yellow circle means the weights of novel categories. The labels of the base (B) and novel (N) categories are shown on the far right. For all experiments, we used SGD with 300 iterations and the learning rates were (a) 1e-2, (b) 1e-4 and (c) 1e-3. \vspace{-2mm}
}\label{fig:3d_vis}
\end{figure*}

One simple way to exploit training examples for a new category is to fine-tune a network by adding nodes to a classifier. However, this approach can destroy a feature space well constructed for base categories owing to the small number of training examples \cite{mao2015learning}. Instead of fine-tuning a network, recent studies have attempted to train it to become generalizable to unseen categories. For example, training examples for base categories can be used to train a weight predictor where the weights of both base and novel categories are calculated using activations from a feature extractor \cite{qiao2017few}. Similarly, the weights of novel categories can be generated by exploiting the weights of base categories with an attention mechanism \cite{gidaris2018dynamic}. To predict the weights of novel categories, these studies used activations or weights well learned for base categories with a large number of training examples. Nevertheless, the feature space for novel categories has not been explicitly considered, and whether the feature space trained for base categories is suitable for novel categories unseen during the training procedure is also uncertain.

In this paper, we propose a fine-tuning strategy that trains novel categories with a few training examples while not contaminating the feature space well learned for the base categories. The goal is to allow the base and novel categories to co-exist in a common space without interference. To this end, we first restrict the feature space to a high-dimensional sphere by normalizing the features and weights. This eliminates information related to magnitudes, and thus makes it easy to control the feature space using only angular information. Based on the high-dimensional sphere, we propose two geometric constraints. The first is called weight-centric feature clustering. For a given category, when used with cross-entropy loss, this reduces the angular distance between the features and the weights. The second is angular weight separation; this separates category weights based on the angular distance.

Fig. \ref{fig:3d_vis} shows our motivation for the proposed geometric constraints. From the CIFAR 100 dataset \cite{krizhevsky2009learning}, we chose 10 base categories consisting of five pairs, each from a parent category. For example, the images of ``Motorcycle'' and ``Bicycle'' were included in the parent category, Vehicles 1. For novel categories, we chose ``Cockroach'' and ``Skunk'' which were not similar to any of the base categories, and ``Shark'' whose images looked similar to those in ``Whale''. Using $500$ training examples for each base category, we first trained a shallow convolutional neural network (CNN) that had three nodes for the feature extractor and the softmax classifier followed by cross-entropy loss. We then extracted features of the examples for the base and novel categories. The extracted features are shown in Fig. \ref{fig:3d_vis}. Features of ``Shark'' are located close to those of ``Whale,'' which were well trained, and features of the other two novel categories are spread out. We can also see that merely fine-tuning the novel categories with five training examples contaminates the feature space that was well learned for the base categories. Furthermore, it is clear that our proposed method not only preserves features of the base categories but also constructs discriminative features of novel categories using only five training examples for each category. In Section \ref{sec:exp}, we show that the proposed method can be applied to more complex datasets such as a subset of ImageNet built for few-shot learning.

To sum up, our contributions in this paper are threefold:
\begin{itemize}
	\item 	Assuming only a few training examples for novel categories, a fine-tuning strategy is proposed to classify both base and novel categories. That is, we do not use any training examples of the base categories to fine-tune a network or for few-shot learning.
	\item	We develop two geometric constraints that help classify the novel categories while maintaining the feature space previously learned for the base categories.
	\item	We demonstrate that the proposed method achieves state-of-the art performance and we analyse it using several CNNs.
\end{itemize}

We discuss the overall model in Section \ref{sec:frameworks}. The details of the geometric constraints are described in Section \ref{sec:loss} and the experimental results are detailed in Section \ref{sec:exp}. We discuss the proposed method and the result in Section \ref{sec:discussion}. Finally, we present the conclusions of this work in Section \ref{sec:conclusion}.

\section{Related Work} \label{sec:relatedwork}
A typical few-shot learning problem does not assume the existence of base categories with a large number of training examples, and its objective is to only classify novel categories with a few training examples. Meta-learning has recently attracted interest in the context of solving this problem \cite{munkhdalai2017meta, qi2017learning, snell2017prototypical, vinyals2016matching, yang2018learning}. In meta-learning, we use meta-training, meta-validation and meta-test datasets. Given a meta-training dataset, we randomly select a few training examples to obtain the parameters of the network and randomly choose test examples to generate a loss and train it. This process is repeated several times to make the network generalizable to unseen categories. Finally, the algorithms are evaluated in the meta-test set by the way they used in the meta-training set. Note that the three meta-datasets have disjoint categories.

Based on the concept of meta-learning, a matching network \cite{vinyals2016matching} was proposed to jointly learn two embedding functions for training examples and a test example. This method uses a deterministic distance metric to compare them. Unlike in a fixed manner, the proposed method in \cite{yang2018learning} learns a relation module to compare the results of the two embedding functions. A prototypical network \cite{snell2017prototypical} learns a prototype for each category, so that the examples discriminatively cluster around the prototypes corresponding to each category. Some studies \cite{cai2018memory, munkhdalai2017meta} developed memory modules that store useful information from training examples and exploited their memories when testing them. On the contrary, the relationship between activations and weights was established by \cite{qi2017learning}. The authors used $l2$ normalization layers to benefit from cosine similarity that renders activations and weights in the final layer symmetric. Thus, the activations of training examples for novel categories can be regarded as weights in the final layer.

However, the purpose of the above studies was to only classify novel categories with a few training examples, and they did not assume the existence of previously-trained base categories. In this respect, the most relevant works to ours are \cite{qiao2017few}, \cite{gidaris2018dynamic}, \cite{wang2018low} and \cite{hariharan2017low}. A weight predictor trained on a large number of training examples for base categories was proposed by \cite{qiao2017few}. Following training, the weight predictor was used to calculate the weights of novel categories based on the activation of a few training examples. An attention-based weight generator \cite{gidaris2018dynamic} was developed to predict the weights of novel categories by exploiting those of the base categories. These two studies take advantage of activations or weights that are well learned for the base categories with a large number of training examples, and do not need to fine-tune a network for novel categories. However, it is unclear whether the feature space trained for the base categories is also suitable for novel categories hitherto unseen. 

In terms of a generative model, Wang \textit{et al.} \cite{wang2018low} generated data that has similar characteristics to the training examples for novel categories. Similarly, Hariharan and Girshick \cite{hariharan2017low} presented an example generation function where transformations learned from base categories are applied to the examples of novel categories. Then, they re-trained a classifier using a large dataset for base categories and the generated examples for novel categories. However, the complexity of this training procedure can be burdensome for few-shot learning. Instead, we propose a fine-tuning strategy that uses only a few training examples for novel categories. Moreover, fine-tuning the network does not affect the feature space learned for the base categories and generates discriminative features for novel categories.

Finally, a common concept exploiting marginal distance was proposed by \cite{largemargin2018}. However, the paper aims to classifying only novel categories without any distractors (e.g., base categories). This means their idea cannot be directly applied to our scenario. Specifically, they show the triplet (or contrastive) loss can be applied to few-shots. But, the losses should form triplets of anchor, positive and negative examples for training. Note that, in our scenario, generating triplets for training incurs very high complexity since we deal with both base and novel categories. Meanwhile, our method only uses a few training examples (not pairs of examples) and we will show our method works for prevalent datasets for few-shot image recognition.

\begin{figure*}
	\begin{center}
		\includegraphics[height=9cm]{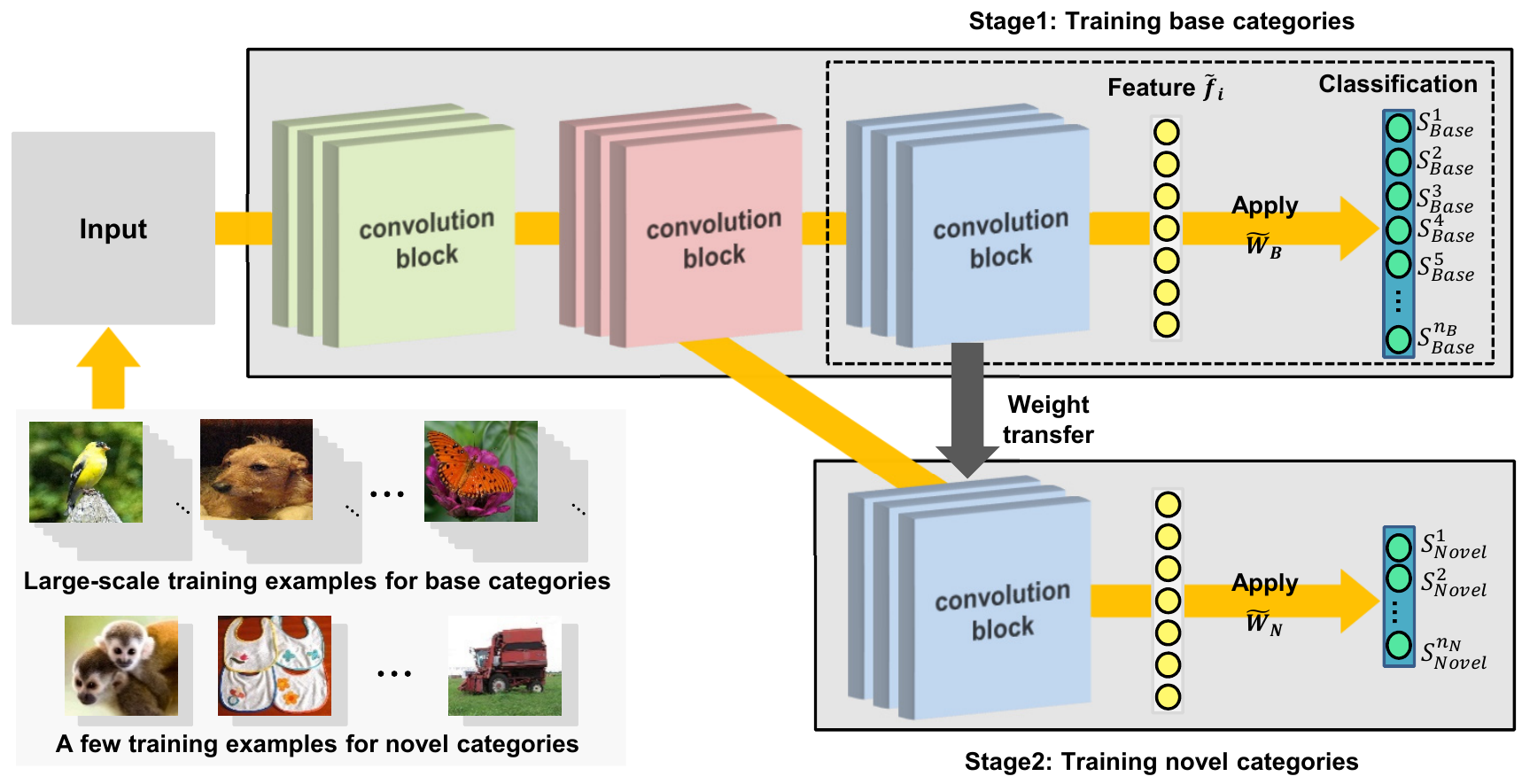}
		\caption{Overall framework. In training stage 1, a network is trained for the base categories with a large number of training examples. In training stage 2, we duplicate parts of the convolution blocks to be fine-tuned and build a classification layer for the novel categories. The parameters of the bottom blocks are frozen. We forward the training examples for the novel categories to the bottom blocks, the duplicated top blocks and the novel classifier. In the test stage, an example is forwarded to all layers, and we calculate the largest value from the classification layers of the base and novel categories.}
		\label{fig:framework}
	\end{center}
\end{figure*}

\section{Overall Framework} \label{sec:frameworks}
In this section, we define the few-shot learning problem, introduce a two-stage training procedure to classify base and novel categories, and explain how to test an example.

\subsection{Problem Definition}
In our classification problem, the dataset $D_{Base}=\{D_{Base}^{Tr}, D_{Base}^{Val}, D_{Base}^{Test}\}$ is available to train, validate, and test the base categories. Similar to ImageNet \cite{russakovsky2015imagenet}, $D_{Base}$ is composed of a large number of examples for $C_{Base}$ categories. Then, (1) we train a network with $D_{Base}^{Tr}$ to classify $C_{Base}$ categories.

Based on the network, we aim to add novel categories to it. The dataset $D_{Novel}$ is available for novel categories, containing $C_{Novel}^{\prime}$ categories that are disjoint to $C_{Base}$. (2) We randomly choose $C_{Novel} \left(\in C_{Novel}^{\prime}\right)$ categories, and $k$ training and $T_{Novel}$ test examples from each category. Then, the network is further trained using $k \times C_{Novel}$ examples. This setting is called $C_{Novel}$-way $k$-shot learning. The typical number of $k$ is one or five.

Finally, (3) the network is evaluated using $T_{Base} \left(\in D_{Base}^{Test}\right)$ and $T_{Novel}$ examples. Thus, given $C_{Both} = C_{Base} \cup C_{Novel}$, the goal is to correctly classify an example as belonging to one of $C_{Both}$ categories. Steps (2) and (3) are repeated several times to obtain a $95\%$ confidence interval. 

\subsection{Two-Stage Training Procedure}
\noindent \textbf{Notation} \quad We consider a network composed of a feature extractor and a classifier. We use $f_i$ to denote the feature extracted from the $i$-th example through the feature extractor. The weights of the base categories for the classifier are $W_B = \left[w_B^1 w_B^2 \cdots w_B^{n_B}\right]$, where $w_B^j$ is the column vector for the $j$-th category and $n_B$ the number of the base categories. Likewise, the weights of the novel categories are denoted by $W_N = \left[w_N^1 w_N^2 \cdots w_N^{n_N}\right]$. If $l2$ normalization is applied, we denote it by $\widetilde{v}={v}/{\Vert v \Vert}$, where $v$ is a column vector and $\widetilde{V} = \left[ \widetilde{v}^1 \widetilde{v}^2 \cdots \widetilde{v}^n \right]$, where $\widetilde{v}^j$ is a normalized column vector. We define scores in a classification layer as $S_{Base}=\{ \widetilde{f}_i^T \widetilde{w}_B^1, \widetilde{f}_i^T \widetilde{w}_B^2, \cdots, \widetilde{f}_i^T \widetilde{w}_B^{n_B}\}$ and $S_{Novel}=\{ \widetilde{f}_i^T \widetilde{w}_N^1, \widetilde{f}_i^T \widetilde{w}_N^2, \cdots, \widetilde{f}_i^T \widetilde{w}_N^{n_N}\}$.
\newline\newline
\noindent \textbf{Training Stage 1} \quad As shown in Fig. \ref{fig:framework}, a network is trained on $D_{Base}^{Tr}$ to classify $C_{Base}$ categories. The features and weights are normalized by $l2$-norm. For classification, we use the softmax layer followed by cross-entropy loss. This can be expressed as
\begin{equation}\label{eqn:loss_cls}
L_{cls} = - \dfrac{1}{M} \sum_{i=1}^{M}  log \dfrac{e^{s \widetilde{f}_i^T \widetilde{w}_B^{y_i} }}{\sum_{j=1}^{n_B} e^{s \widetilde{f}_i^T \widetilde{w}_B^{j}}},
\end{equation}
where $M$ is a batch size, $y_i$ refers to the category of the $i$-th example, and $\widetilde{f}_i^T \widetilde{w}_B^{j}$ is the cosine similarity between the weight and the feature. $s$ is a learnable parameter.

Eq. \ref{eqn:loss_cls} has two notable aspects. First, we normalize the features and weights by $l2$-norm. This locates the feature space in a high-dimensional sphere. In other words, we consider only angular information with respect to features and weights. This helps us in training stage 2 to geometrically control the locations of the features and weights in terms of angular distances. Second, we apply a scale parameter $s$. As addressed in \cite{wang2017normface}, the range $[-1, 1]$ of $\widetilde{f}_i^T \widetilde{w}_B^j$ is too small to obtain a sufficient gradient for training. Thus, it is possible for the network to fail to converge. To solve this problem, we scale up the cosine similarity by using the learnable parameter $s$ \cite{wang2017normface}.

In addition to Eq. \ref{eqn:loss_cls}, we also apply a loss called weight-centric feature clustering that will be introduced in Section \ref{sec:loss}. This is to place the weights for base categories at predictable locations, such as the center of features for each category.
\newline\newline
\noindent \textbf{Training Stage 2} \quad  We now consider $D_{Novel}$ and the network trained in training stage 1. The parameters of bottom blocks are frozen. However, we duplicate the top convolution blocks in order to fine-tune them, and build a classifier consisting of $C_{Novel}$ nodes. Given a training example of $D_{Novel}$, it is forwarded to the bottom blocks, the duplicated top layers, and the novel classifier. In training stage 2, we do not use the top convolution blocks and the classifier used for training the base categories, which are shown using broken lines in Fig. \ref{fig:framework}. Based on this network architecture, we use the loss functions proposed in Section \ref{sec:loss} to train examples of novel categories.

\subsection{Two-Stream Test Procedure}
Once training stages 1 and 2 are complete, we use test examples from $D_{Base}^{Te}$ and $D_{Novel}$. We forward a test example $I_i$ to all layers used in training stages 1 and 2. The example is then classified into a category by
\begin{equation}
\underset{k}{\mathrm{argmax}}(S_{Both}^k),
\end{equation}
where $S_{Both}^k$ is the $k$-th element of $S_{Base} \cup S_{Novel}$. To evaluate base categories or novel categories only, we substitute $S_{Both}^k$ into $S_{Base}^k$ or $S_{Novel}^k$, respectively.

\begin{figure*}
		\centering
        \subfigure[Feature space after training stage 1]
        {
                \centering
                \includegraphics[width=.43\linewidth]{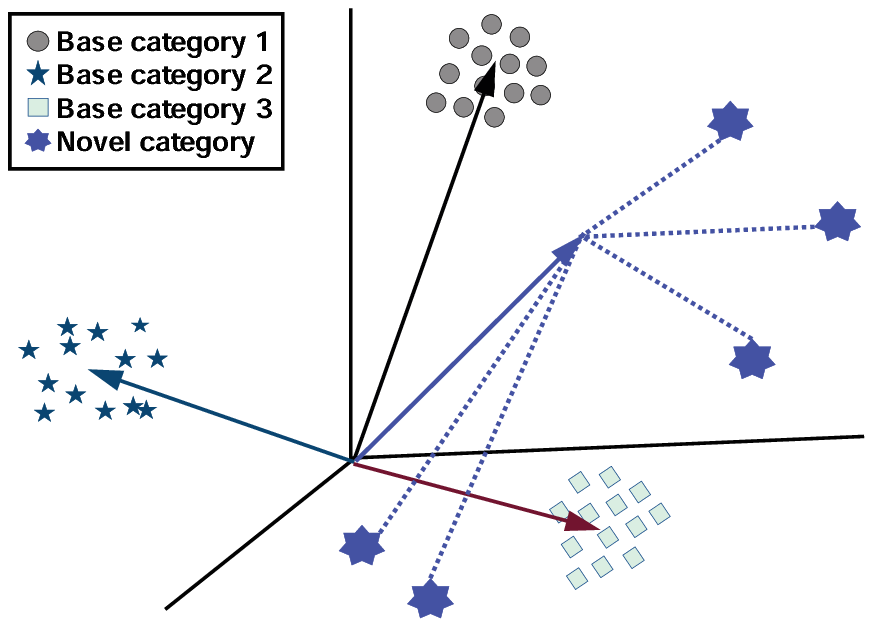}
                \label{fig:before_tr}
		}
        \subfigure[Feature space after training stage 2]
        {
                \centering
                \includegraphics[width=.43\linewidth]{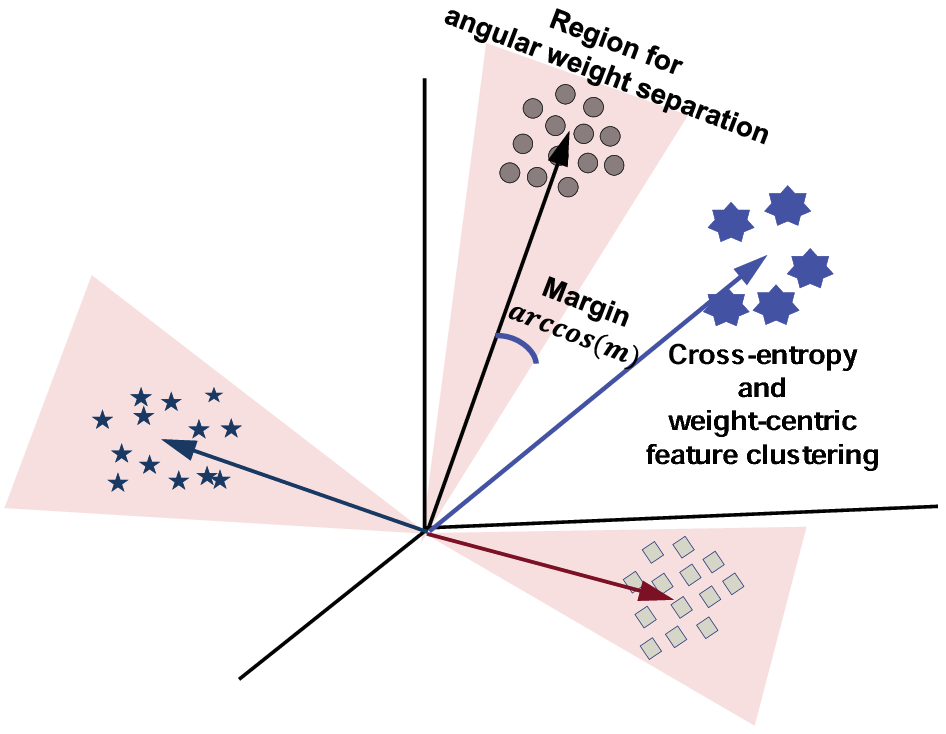}
                \label{fig:after_tr}
		}
        \caption{Visual interpretation of training stages 1 and 2 in 3-dimensional feature space. (a) After the base categories have been trained, their features are clustered close to the weights. (b) By applying the fine-tuning strategy using the proposed loss functions, features of the novel categories are clustered near the weights, and are sufficiently distant from weights of other categories. Note that all features and weights are located on a high-dimensional sphere in practice.}\label{fig:loss_vis}
\end{figure*}

\section{Loss Functions} \label{sec:loss}
In this section, we elaborate on the loss functions used in training stage 2. As we use separate blocks and weights learned \textit{independently} for the base and novel categories, it is unnatural for them to co-exist in a common feature space. To address this issue, we introduce three functions, cross-entropy loss and two geometric constraints, to extract discriminative features for novel categories while preserving the feature space trained for the base categories in training stage 1.

\subsection{Cross-Entropy Loss}
We use a similar loss function to Eq. \ref{eqn:loss_cls} but replace the weights for the base categories with those for the base and novel categories and use the fixed scale parameter $s$ trained in training stage 1.

\subsection{Weight-Centric Feature Clustering}
Because novel categories are not trained in stage 1, their features may not be well clustered, and thus intra-class variation may be large. To solve this problem, we propose the weight-centric feature clustering (WCFC) defined as


\begin{equation} \label{eqn:WCFC}
L_{WCFC} = \sum_{i=1}^{C_{Novel}} -log (cos \ \theta_{{g({f}^i), \widetilde{w}_N^i}}),
\end{equation}
where $cos \ \theta_{{g({f}^i),\widetilde{w}_N^i}} = g({f}^i)^T \cdot \dfrac{w_N^i}{\Vert w_N^i \Vert}$. We define a function g($\cdot$) as two types. For type 1,

\begin{equation} \label{eqn:WCFC_1}
	\qquad g({f}^i) = \dfrac{\bar{f}^i}{\Vert \bar{f}^i \Vert},
\end{equation}
where $\bar{f}^i$ is the arithmetic mean of features of the training examples for the $i$-th category. As the initial weight $w_N^i$, we use $\bar{f}^i$. For type 2,

\begin{equation} \label{eqn:WCFC_2}
	\qquad g({f}^i) = \dfrac{\sum_i \widetilde{f}^i}{\Vert \sum_i \widetilde{f}^i \Vert}.
\end{equation}
When using this type, we set the initial weight $w_N^i$ to $\sum_i \widetilde{f}^i$.
 
Note that Eq. \ref{eqn:WCFC_1} is the normalization of averaged features and Eq. \ref{eqn:WCFC_2} is the normalization of the sum of normalized features. Thus, Eq. \ref{eqn:WCFC_2} has more degrees of freedom for magnitude and we have found that this is beneficial to novel categories for few-shot learning.

When used with cross-entropy loss, this weight-centric feature clustering states that features corresponding to a category should be located near the weight. In Eq. \ref{eqn:WCFC}, maximizing the cosine similarity $cos \theta_{{g({f}^i),\widetilde{w}_N^i}}$ is the core concept.

\subsection{Angular Weight Separation}
Thus far, we have clustered features for novel categories near the weights to reduce intra-class variation. Our goal now is to ensure that the weights and features for the novel categories are sufficiently far from those for the base categories. To achieve this, we first define the angular distance $u_{ij}$ between weights as follows:
\begin{equation}\label{eqn:weight_distance}
	u_{ij} =
		\begin{cases}
			cos \ \theta_{\widetilde{w}^i, \widetilde{w}^j_N}, & \text{if} \ 
									\widetilde{w}^i \not\equiv \widetilde{w}^j_N \\
			0, & \text{otherwise}
		\end{cases},	
\end{equation}
where $\widetilde{w}^i$ is the $i$-th column vector of $\left[ \widetilde{W}_B \widetilde{W}_N \right]$.

Then, we maximize $\theta_{\widetilde{w}^i, \widetilde{w}^j_N}$, so that $u_{ij}$ is smaller than margin $m$. This is called the angular weight separation (AWS) constraint and is defined as
\begin{equation}\label{eqn:AWS}
L_{AWS} = \dfrac{\sum_{i,j} -log \left( - u_{ij} \cdot \mathbbm{1}_M \left( u_{ij} \right) + 1 \right)}{\sum_{i,j} \mathbbm{1}_M \left( u_{ij} \right)},
\end{equation}
where $M = \left\lbrace  u_{ij} \ | \ u_{ij} > m, \ \forall i, j \right\rbrace$ and 
$\mathbbm{1}_M \left( u_{ij} \right) = 
	\begin{cases}
		1, & \text{if} \ u_{ij} \in M \\
		0, & \text{otherwise}
	\end{cases}
$ is the indicator function. The logarithm and a factor of $1$ are used to align the loss function with the scale of the other loss functions. During the training process, when all novel weights are separated from other weights in terms of the angular distance, we turn off the AWS loss.

\subsection{Regularization}\label{subsec:reg}
We sum up all the losses for the final loss function.
\begin{equation} \label{eqn:tot}
	L_{total} = \gamma L_{cls} + \alpha L_{WCFC} + \beta L_{AWS}
\end{equation}
Using Eq. \ref{eqn:tot}, we train the convolution blocks and weights for the novel categories in training stage 2. Fig. \ref{fig:loss_vis} shows  the role of each loss function. For all experiments, we use $\gamma\alpha\beta=111$. In Section V, we discuss how different settings of the hyper-parameters impact on the performance of the proposed method.

\section{Experiments} \label{sec:exp}
To evaluate the proposed method, we used two common datasets for few-shot learning: \textit{mini}ImageNet \cite{vinyals2016matching} and Bharath $\&$ Girshick's dataset \cite{hariharan2017low}. The two datasets were built based on ImageNet \cite{russakovsky2015imagenet}.

\subsection{\textit{mini}ImageNet}
\textit{Mini}ImageNet \cite{vinyals2016matching} is the most commonly used dataset for few-shot learning. It is composed of $64$ training, $16$ validation, and $20$ test categories. Each category contains $600$ examples of size $84 \times 84$. We used the split provided by \cite{ravi2016optimization}. Most studies use \textit{mini}ImageNet as follows: Using the training categories, $k$ training examples for each of $C_{Novel}$ categories are chosen to obtain the parameters of a network and $T$ test examples are randomly chosen to train it. Then, this process is repeated several times to warm-up or generalize the parameters of the network. Finally, the network is evaluated on the test categories in the same way as it was on the training categories. Therefore, this dataset has been primarily used to only classify novel categories in meta-learning.

\setlength\dashlinedash{1pt}
\setlength\dashlinegap{1.5pt}
\setlength\arrayrulewidth{0.3pt}
\begin{table*}[h!]
\caption{Classification accuracy on \textit{mini}ImageNet. The results are averaged to obtain a 95\% confidence interval. $\lq$-' denotes that the performance was not reported for the model. We referred to \cite{gidaris2018dynamic} for the accuracies of other methods. $^\dagger$ indicates that the proposed data augmentation technique was used. The highest numbers are bolded.}\label{table:mini}

\centering
 \resizebox{\textwidth}{!}{
 \renewcommand{\arraystretch}{0.9}
    \tiny
    \begin{tabular}{c|c|ccc|ccc}
    \hline
    \multirow{2}[2]{*}{Models} & \multirow{2}[2]{*}{Feature Extractors} & \multicolumn{3}{c|}{5-way 5-shot}                         & \multicolumn{3}{c}{5-way 1-shot} \\
                      &                   & Novel             & Both              & Base              & Novel             & Both              & Base \bigstrut[b]\\
    \hline
    Matching Nets \cite{vinyals2016matching} & C64F              & 55.30             & -                 & -                 & 43.60             & -                 & - \bigstrut[t]\\
    Finn \textit{et al.} \cite{finn2017model}  & C64F              & 63.10$\pm$0.92        & -                 & -                 & 48.70$\pm$1.84        & -                 & - \\
    Prototypical Nets \cite{snell2017prototypical} & C64F              & 68.20$\pm$0.66        & -                 & -                 & 49.42$\pm$0.78        & -                 & - \\
    Relation Nets \cite{yang2018learning} & C64F              & 65.32$\pm$0.66        & -                 & -                 & 50.44$\pm$0.82        & -                 & - \\
    Mishra \textit{et al.} \cite{mishra2018simple} & ResNetS           & 68.88$\pm$0.92        & -                 & -                 & 55.71$\pm$0.99        & -                 & - \bigstrut[b]\\
    \hline
    \multirow{2}[2]{*}{Gidaris and Komodakis  \cite{gidaris2018dynamic}$^\dagger$} 
    & C64F              
    & 73.27$\pm$0.59        & 57.72             & 66.96             
    & 57.78$\pm$0.78        & 48.13             & 66.58 \bigstrut[t]\\
                      
	& ResNetS           
	& 70.32$\pm$0.66        & 55.93             & 79.58            
	& 56.76$\pm$0.80        & 49.68             & 79.43 \bigstrut[b]\\
    \hline

    \multirow{3}[2]{*}{Proposed} 
    & C64F              
    & 72.92$\pm$0.64        & 58.00             & 67.12             
    & 53.39$\pm$0.81        & 45.32             & 67.12 \bigstrut[t]\\
                      
    & C64F-Dropout      
    & 73.21$\pm$0.65        & 58.77             & 66.98             
    & 55.74$\pm$0.83        & 48.11             & 66.98 \\
                      
    & \textbf{ResNetS}  
    & \textbf{78.00$\pm$0.61} & \textbf{68.16}    & \textbf{79.78}             
    & \textbf{58.52$\pm$0.82} & \textbf{56.05}    & \textbf{79.78} \bigstrut[b]\\
    
    \hdashline
    \multirow{3}[2]{*}{Ablation (w/o Fine-Tuning)} 
    & C64F              
    & 72.85$\pm$0.63        & 54.43             & 67.12             
    & 53.43$\pm$0.81        & 43.17             & 67.12 \bigstrut[t]\\
    
    & C64F-Dropout      
    & 71.32$\pm$0.63        & 51.71             & 66.98             
    & 54.86$\pm$0.80        & 38.52             & 66.98 \\
                      
    & ResNetS           
    & 68.68$\pm$0.67        & 47.11             & 79.78             
    & 54.12$\pm$0.83        & 43.11             & 79.78 \bigstrut[b]\\
    \hline
    \end{tabular}%
}
\end{table*}

Recently, Gidaris and Komodakis \cite{gidaris2018dynamic} extended the dataset by collecting $300$ extra validation examples and $300$ test examples for the $64$ base categories. Thus, we used the original $600$ examples to train the base categories and the additional examples for validation and testing. We trained and tested the base and novel categories as follows: In training stage 1, we trained $64$ base categories with $600$ examples. In training stage 2, we randomly chose $C_{Novel}$ novel categories and $k$ examples for each category to fine-tune the network using the proposed loss functions. In the test stage, we randomly chose $T$ examples from each of the $64$ base categories and $C_{Novel}$ novel categories to test the network. We iterated training stage 2 and the test stage to obtain a $95\%$ confidence interval.

For the feature extractor, we used a shallow CNN called C64F with four convolution blocks: each block had $3 \times 3$ convolutions, batch normalization \cite{ioffe2015batch}, ReLU \cite{nair2010rectified} and $2 \times 2$ max-pooling. The sizes of feature maps for the four convolution blocks were $64$. For a deeper model, the small version of ResNet \cite{he2016deep} proposed in \cite{mishra2018simple} was used as in \cite{gidaris2018dynamic}. We call the network ResNetS. In all cases, we did not use ReLU for the last layer in the last convolutional block.
\newline\newline
\noindent \textbf{Data Augmentation} \quad As our method uses a fine-tuning strategy, it is important for training examples to generate a loss large enough to back-propagate the error. In this respect, we found that augmenting the training examples of each novel category helped improve performance. Given a training example, we performed zero-padding with 8 pixels on each border and randomly cropped it to make $84 \times 84$. It was then horizontally flipped with a probability of $0.5$. This transformation is identical to that used when training the base categories in training stage 1. By iterating this process, we generated $nAug$ examples for each training example. In our experiments, we augmented examples to have $20$ training examples for each novel category in total. Since this simple augmentation helps improve performance, more advanced example generation methods \cite{hariharan2017low, radford2015unsupervised} may further boost the performance of our algorithm.
\newline\newline
\noindent \textbf{Effect of Dropout} \quad Since our algorithm uses a fine-tuning strategy, it is crucial that a network has an ability to be generalizable to unseen categories. Typically, we rely on a large number of training examples to train a network with a highly complicated distribution. For few-shot learning, however, we only have a few training examples, and thus other techniques for generalization should be considered. In our experiment, we have found that ResNetS significantly outperforms C64F. We assume that this is because not only the depth of the network but also the usage of Dropout \cite{srivastava14a} on the last fully connected layers. Thus, we add Dropout following a linear layer with $1,024$ neurons to C64F and set the dropout rate to 0.5. We call the network C64F-Dropout.

\begin{table*}[]
\caption{Top-5 classification accuracy on the ImageNet. The results are averaged to obtain a 95\% confidence interval. $\lq$-' denotes that the performance was not reported for the model. $\lq$w/ H' is short for w/ Hallucination where the algorithms use example generation techniques.  $^\dagger$ indicates that the proposed data augmentation technique was used. The highest numbers are bolded. *Results were reported by \cite{wang2018low} and we referred to \cite{gidaris2018dynamic}.}\label{table:large}
\centering
\resizebox{\textwidth}{!}{
\renewcommand{\arraystretch}{1.2}
\Huge
    \begin{tabular}{c|ccccc|cccccc|cccccc}
    \hline
    \multirow{2}[2]{*}{Models} & \multicolumn{5}{c|}{Novel}                                                                        &                   & \multicolumn{5}{c|}{Both}                                                                          &                   & \multicolumn{5}{c}{Both with prior} \bigstrut[t]\\
                      & k=1               & 2                 & 5                 & 10                & 20                &                   & k=1               & 2                 & 5                 & 10                & 20                &                   & k=1               & 2                 & 5                 & 10                & 20 \bigstrut[b]\\
    \hline
    Prototypical Nets \cite{snell2017prototypical}* & 39.3              & 54.4              & 66.3              & 71.2              & 73.9              &                   & 49.5              & 61                & 69.7              & 72.9              & 74.6              &                   & 53.6              & 61.4              & 68.8              & 72                & 73.8 \bigstrut[t]\\
    Matching Nets \cite{vinyals2016matching}* & 43.6              & 54                & 66                & 72.5              & 76.9              &                   & 54.4              & 61                & 69                & 73.7              & 76.5              &                   & 54.5              & 60.7              & 68.2              & 72.6              & 75.6 \\
    Logistic Regression \cite{wang2018low}* & 38.4              & 51.1              & 64.8              & 71.6              & 76.6              &                   & 40.8              & 49.9              & 64.2              & 71.9              & 76.9              &                   & 52.9              & 60.4              & 68.6              & 72.9              & 76.3 \\
    Logistic Regression w/ H \cite{hariharan2017low}* & 40.7              & 50.8              & 62                & 69.3              & 76.5              &                   & 52.2              & 59.4              & 67.6              & 72.8              & 76.9              &                   & 53.2              & 59.1              & 66.8              & 71.7              & 76.3 \\
    SGM w/ H \cite{hariharan2017low} & -                 & -                 & -                 & -                 & -                 &                   & 54.3              & 62.1              & 71.3              & 75.8              & 78.1              &                   & -                 & -                 & -                 & -                 & - \\
    Batch SGM \cite{hariharan2017low} & -                 & -                 & -                 & -                 & -                 &                   & 49.3              & 60.5              & 71.4              & 75.8              & 78.5              &                   & -                 & -                 & -                 & -                 & - \\
    Prototype Matching Nets w/ H \cite{wang2018low}* & 45.8              & 57.8              & 69                & 74.3              & 77.4              &                   & 57.6              & 64.7              & 71.9              & 75.2              & 77.5              &                   & 56.4              & 63.3              & 70.6              & 74                & 76.2 \\
    Prototype Matching Nets \cite{wang2018low}* & 43.3              & 55.7              & 68.4              & 74                & 77                &                   & 55.8              & 63.1              & 71.1              & 75                & 77.1              &                   & 54.7              & 62                & 70.2              & 73.9              & 75.9 \\

\multirow{2}[1]{*}{Gidaris and Komodakis  \cite{gidaris2018dynamic}$^\dagger$} 
    
& 49.08             & 59.97             & 70.77             & 75.51     	& 78.26  &                   & 60.03             & 66.71             & 73.77             & 76.94			& 78.77  &                   & 58.65             & 65.3              & 72.37				& 75.54         & 77.34 \\
                      
& $\pm$.23          & $\pm$.15          & $\pm$.10          & $\pm$.06      & $\pm$.05 &                   & $\pm$.14          & $\pm$.10          & $\pm$.06          & $\pm$.04 		& $\pm$.03 &                   & $\pm$.14          & $\pm$.09          & $\pm$.06			& $\pm$.04      & $\pm$.03 \bigstrut[b]\\

\hline
\multirow{2}[1]{*}{Proposed} 
& 45.98             & 58.37             & 70.17             & 75.49         & 78.62    &                   & 57.11             & 64.53             & 71.77             & 75.20         & 78.10    &                   & 56.24             & 63.77             & 71.43             & 74.91         & 77.08 \bigstrut[t]\\
& $\pm$.22          & $\pm$.16          & $\pm$.09          & $\pm$.07      & $\pm$.05 &                   & $\pm$.14          & $\pm$.10          & $\pm$.07          & $\pm$.06      & $\pm$.03 &                   & $\pm$.14          & $\pm$.10          & $\pm$.06          & $\pm$.05      & $\pm$.03 \\

\multirow{2}[1]{*}{Proposed-Dropout} 
& \textbf{49.57}    & \textbf{60.89}    & \textbf{71.07}	& \textbf{76.11}			& \textbf{78.84}    &                   & \textbf{60.39}             & \textbf{67.44}				& \textbf{74.22}				& \textbf{77.32}         & \textbf{79.05}    &                   & \textbf{58.91}				& \textbf{65.84}				& \textbf{72.71}             & \textbf{75.96}			& \textbf{77.88} \\
& \textbf{$\pm$.24}			& \textbf{$\pm$.16}			& \textbf{$\pm$.09}			& \textbf{$\pm$.07}		& \textbf{$\pm$.05} &                   & \textbf{$\pm$.14}			& \textbf{$\pm$.10}			& \textbf{$\pm$.06}			& \textbf{$\pm$.05}      & \textbf{$\pm$.03} &                   & \textbf{$\pm$.14}			& \textbf{$\pm$.10}			& \textbf{$\pm$.06}          & \textbf{$\pm$.05}		& \textbf{$\pm$.04} \bigstrut[b]\\
    \hline
    \end{tabular}}
\end{table*}

\noindent \textbf{Performance} \quad In Table \ref{table:mini}, we compare our proposed method with prevalent methods \cite{finn2017model, mishra2018simple, snell2017prototypical, vinyals2016matching, yang2018learning} for few-shot learning and a model \cite{gidaris2018dynamic} considering both base and novel categories. It is clear that, as the feature extractor has a greater capacity to handle the training dataset, the performance of the proposed method improves significantly. For example, in training stage 1, ResNetS yields the accuracy of around $79\%$ in the base categories, higher than the $65\sim68\%$ of C64F. In this case, our method outperforms others by a large margin. Therefore, the proposed fine-tuning strategy with geometric constraints is clearly advantageous when a deeper network is available. It is worth noting that the proposed strategy dose not affect the accuracy on base categories even after fine-tuning the network. For ResNetS, the performance improvements in both categories indicate that the geometric constraints not only preserve the feature space of base categories, but also extract discriminative features for novel categories, when compared to \cite{gidaris2018dynamic} that aims to obtain the weights of novel categories based on those of base categories. Note that Table \ref{table:mini} results from the choice of the weight-centric feature clustering as follows: we always use Eq. \ref{eqn:WCFC_2} for training stage 2. For training stage 1, we used Eq. \ref{eqn:WCFC_2} for C64F; and Eq. \ref{eqn:WCFC_1} for ResNetS. We used $0.6$ for the margin value for the angular weight separation.
\newline\newline
\noindent \textbf{Ablation Study} \quad To assess the benefit of our fine-tuning strategy, we show the accuracy when not fine-tuning the network. In this case, the weights of novel categories are set to the sum of normalized features of training examples for each category. The weights and the feature extractor for novel categories are not trained further. We observe that our fine-tuning strategy significantly improves the performance when evaluating both the base and novel categories.

\subsection{Bharath $\&$ Girshick's Dataset}
Bharath $\&$ Girshick proposed a larger dataset for few-shot learning \cite{hariharan2017low}. The dataset is composed of 193 base categories and 300 novel categories for cross-validation, and there are 196 base categories and 311 novel categories for evaluation. The categories have training, validation and test examples as in ImageNet \cite{russakovsky2015imagenet}. We used the categories provided by the most recent work \cite{gidaris2018dynamic} and we show the performance using the evaluation categories. As a feature extractor, we exploit ResNet10 \cite{he2016deep}. We further show the performance can be boosted with ResNet10-Dropout that adds Dropout \cite{srivastava14a} of the dropout rate 0.2 following a linear layer with $1,024$ neurons. 
\newline\newline
\noindent \textbf{Performance} \quad For this dataset, we provide the performance called Both with prior \cite{wang2018low}. When computing a probability that an image $x$ belongs to a class $k$, $p_k(x)=p(y=k|x)$, this considers a prior probability of whether an example belongs to base categories or novel categories. By following \cite{gidaris2018dynamic}, we set $p(y \in C_{base} | x)=0.2$ and $p(y \in C_{novel} | x)=0.8$ for Both with prior. Table \ref{table:large} shows top-5 classification accuracy on the ImageNet. Our algorithm outperforms the state-of-the art methods for all shots. It is worth noting that without any hallucination and the training examples of base categories\cite{hariharan2017low, wang2018low}, our algorithm improves the performance only using a few training examples for novel categories. The result also verifies that a fine-tuning strategy for few-shot learning can be boosted using a network with Dropout. We used $0.4$ for the margin value for the angular weight separation.

\subsection{Incremental Learning}
Although the literature of few-shot learning mainly considers the performance when $k$-shot training examples are available, we would like to provide a possible application of the proposed method. Since our method fine-tunes a network, one might be interested in the performance when training examples are given gradually. For this experiment, we fine-tuned the feature extractor for novel categories continually from 1-shot to 20-shots with ResNet10-Dropout. As shown in Table \ref{table:inc_learning}, the performance of incremental learning is superior to 20-shots. This suggests that a recognition system can be continually upgraded by the proposed method as novel examples are continually provided.

\setlength\dashlinedash{1pt}
\setlength\dashlinegap{1.5pt}
\setlength\arrayrulewidth{0.3pt}
\begin{table}
\caption{Classification accuracy on the ImageNet using incremental learning. For the proposed-dropout, we used 20-shots and for incremental learning, we fine-tuned the feature extractor gradually from 1-shot to 20-shots.}\label{table:inc_learning}

\centering
\resizebox{\columnwidth}{!}{
 \renewcommand{\arraystretch}{1.2}
    \large
    \begin{tabular}{c|c|c|c}
    \hline
    Models            & Novel             & Both              & Both with prior \bigstrut\\
    \hline
    Proposed-Dropout  & 78.84$\pm$.05             & 79.05$\pm$.03             & 77.88$\pm$.04 \bigstrut[t]\\
    \hline
    Incremental learning & 79.87$\pm$.05             & 79.44$\pm$.04             & 78.59$\pm$.04 \bigstrut[b]\\
    \hline
    \end{tabular}%
}
\end{table}

\begin{figure*}
	\centering
	\subfigure[2D embedding space before fine-tuning]
	{
		\centering
		\includegraphics[width=.42\linewidth, height=7cm]{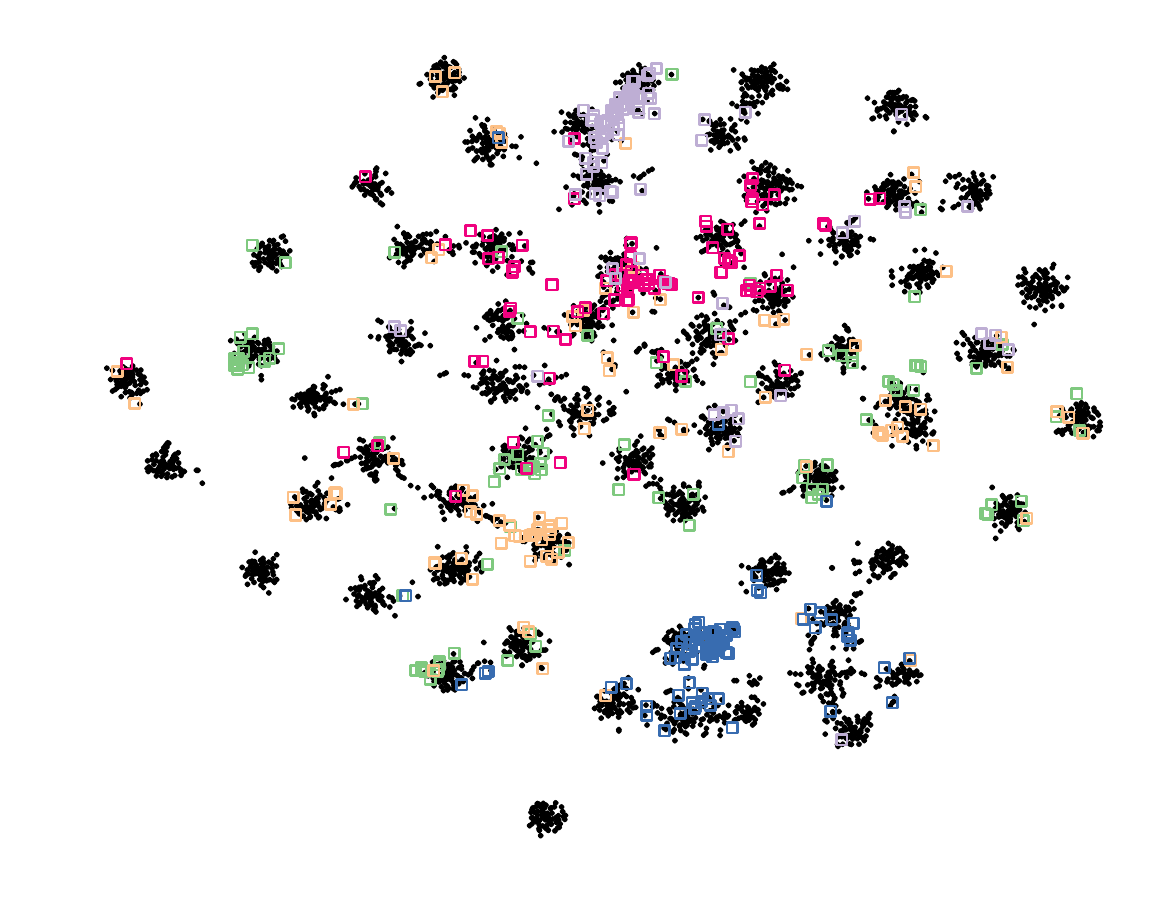}
	} \hspace{2cm}
	\subfigure[2D embedding space after fine-tuning]
	{
		\centering
		\includegraphics[width=.42\linewidth, height=7cm]{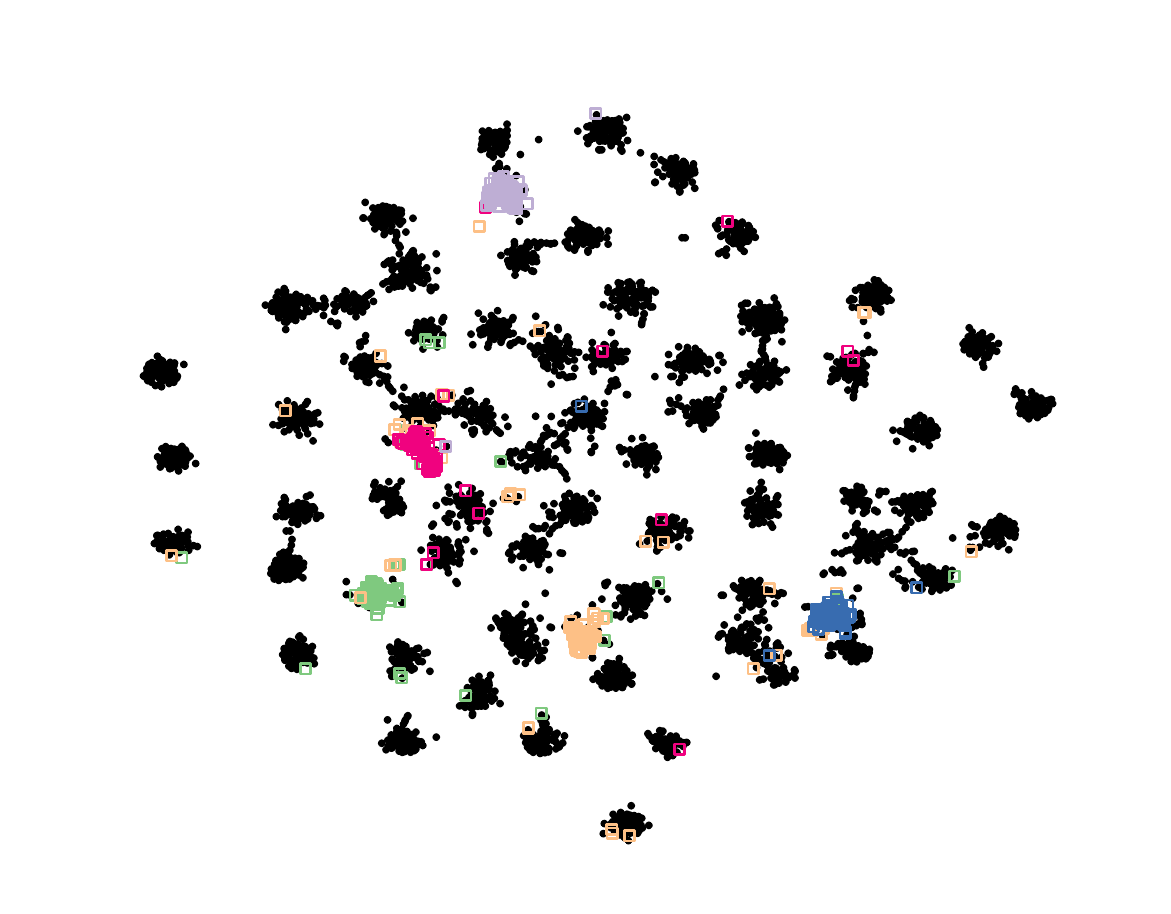}
	}
	
	\subfigure[Novel category 1]
	{
		\centering
		\includegraphics[width=.18\linewidth]{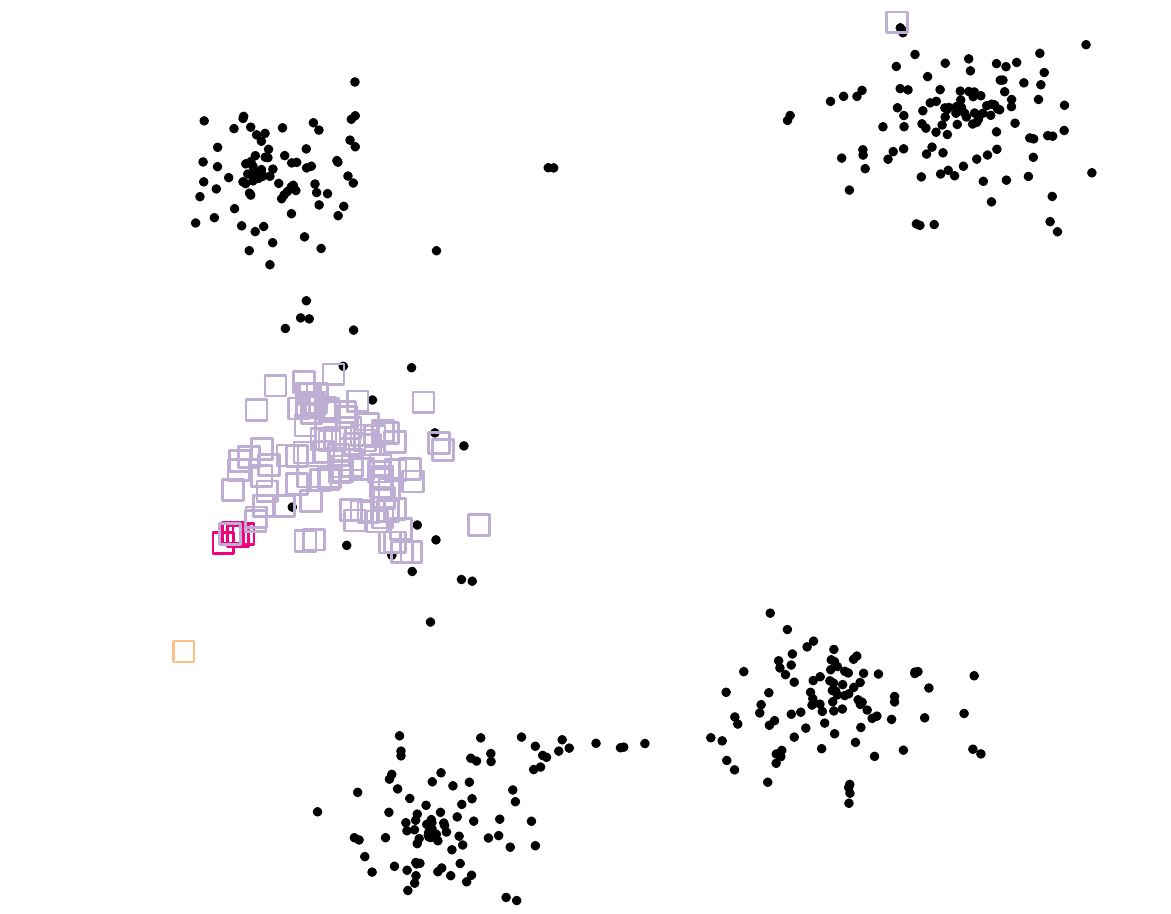}
	}
	\subfigure[Novel category 2]
	{
		\centering
		\includegraphics[width=.18\linewidth]{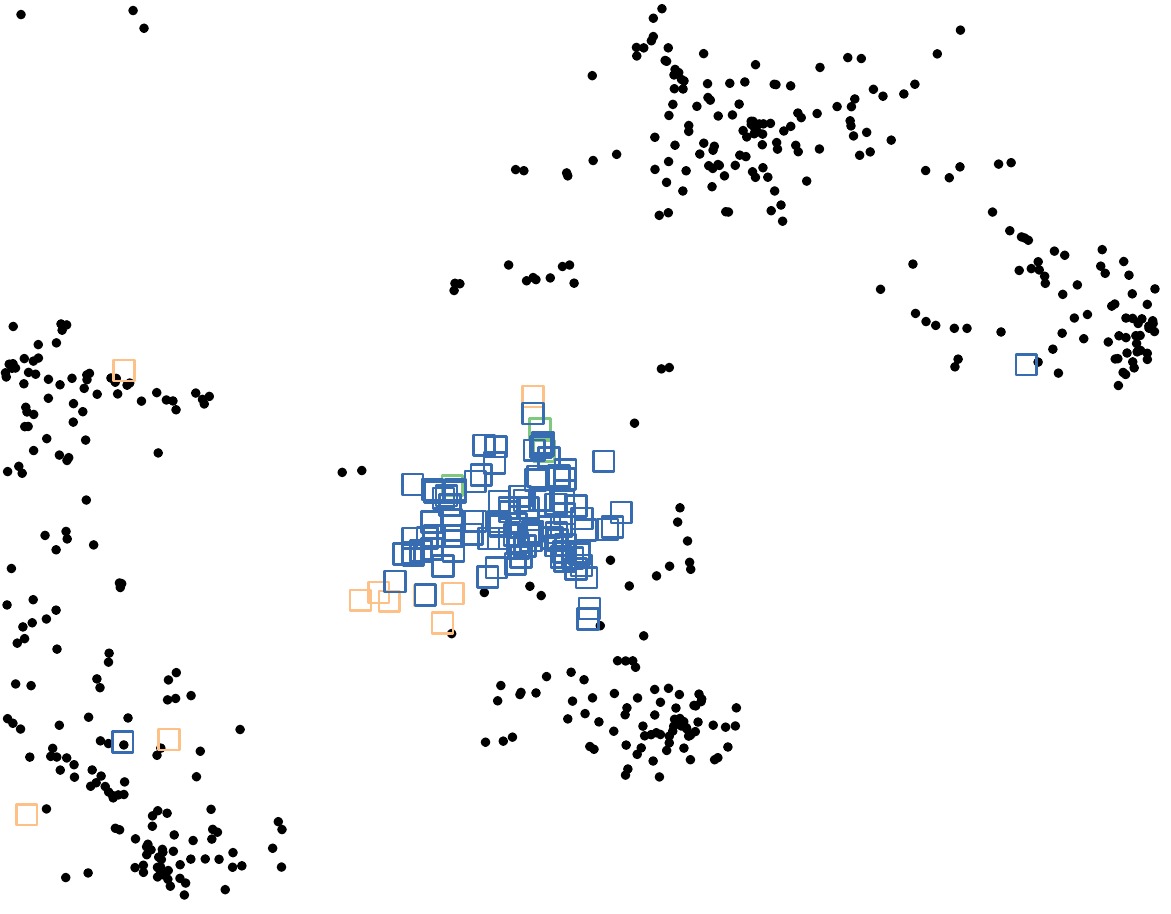}
	}
	\subfigure[Novel category 3]
	{
		\centering
		\includegraphics[width=.18\linewidth]{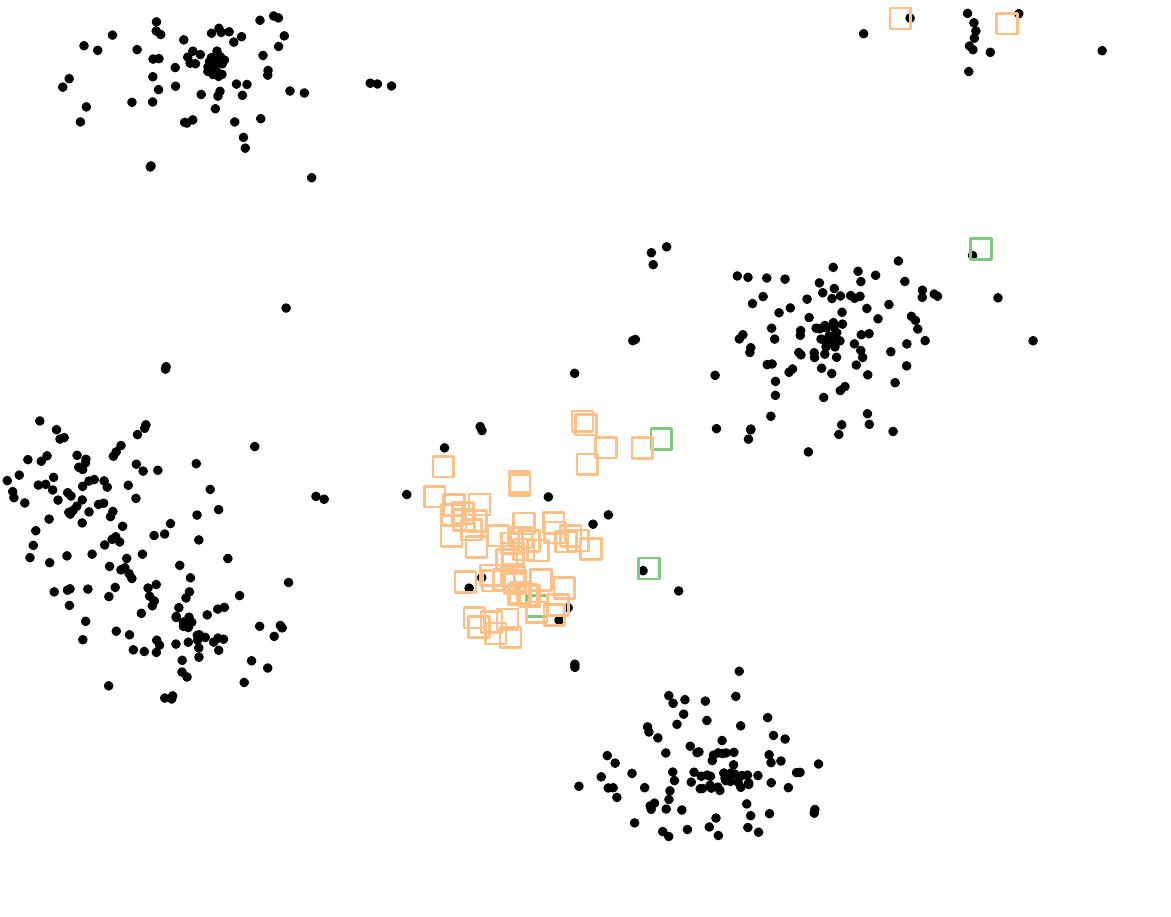}
	}
	\subfigure[Novel category 4]
	{
		\centering
		\includegraphics[width=.18\linewidth]{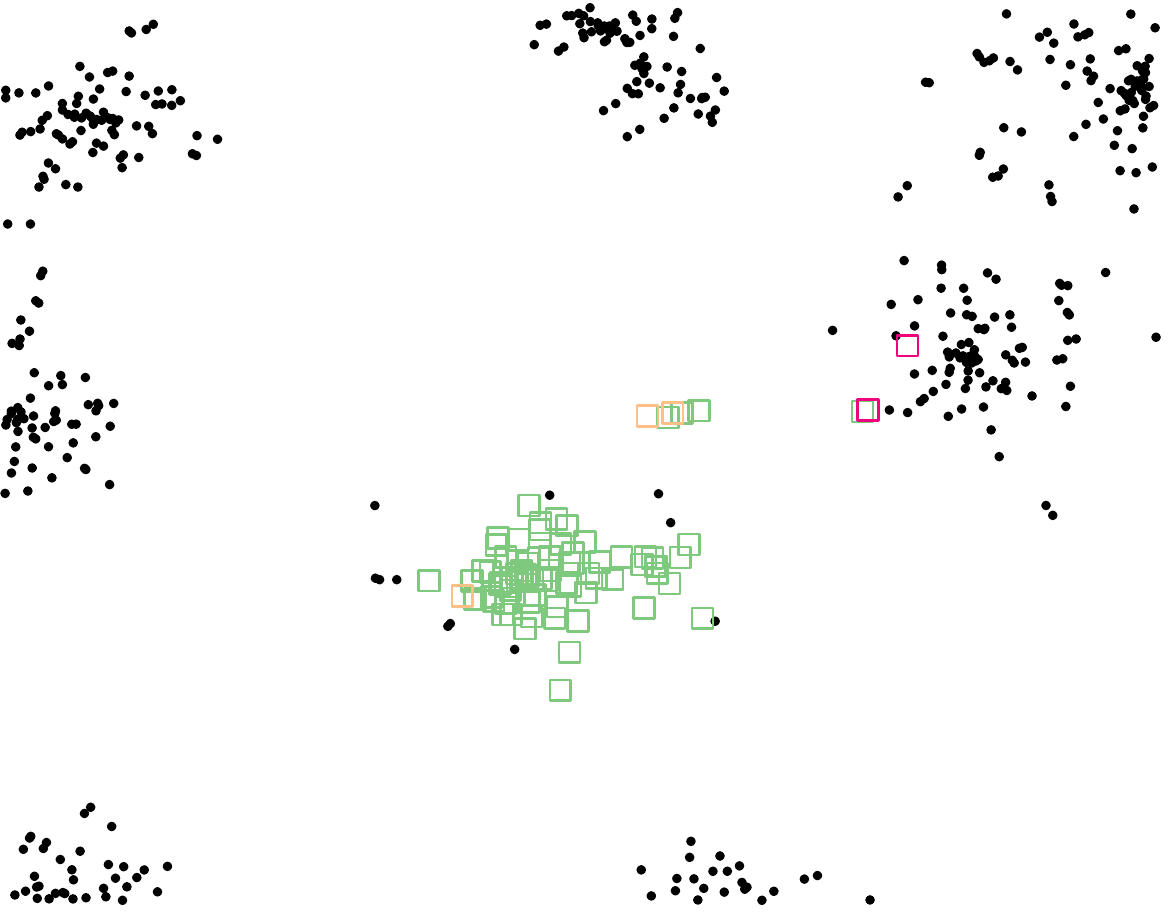}
	}
	\subfigure[Novel category 5]
	{
		\centering
		\includegraphics[width=.18\linewidth]{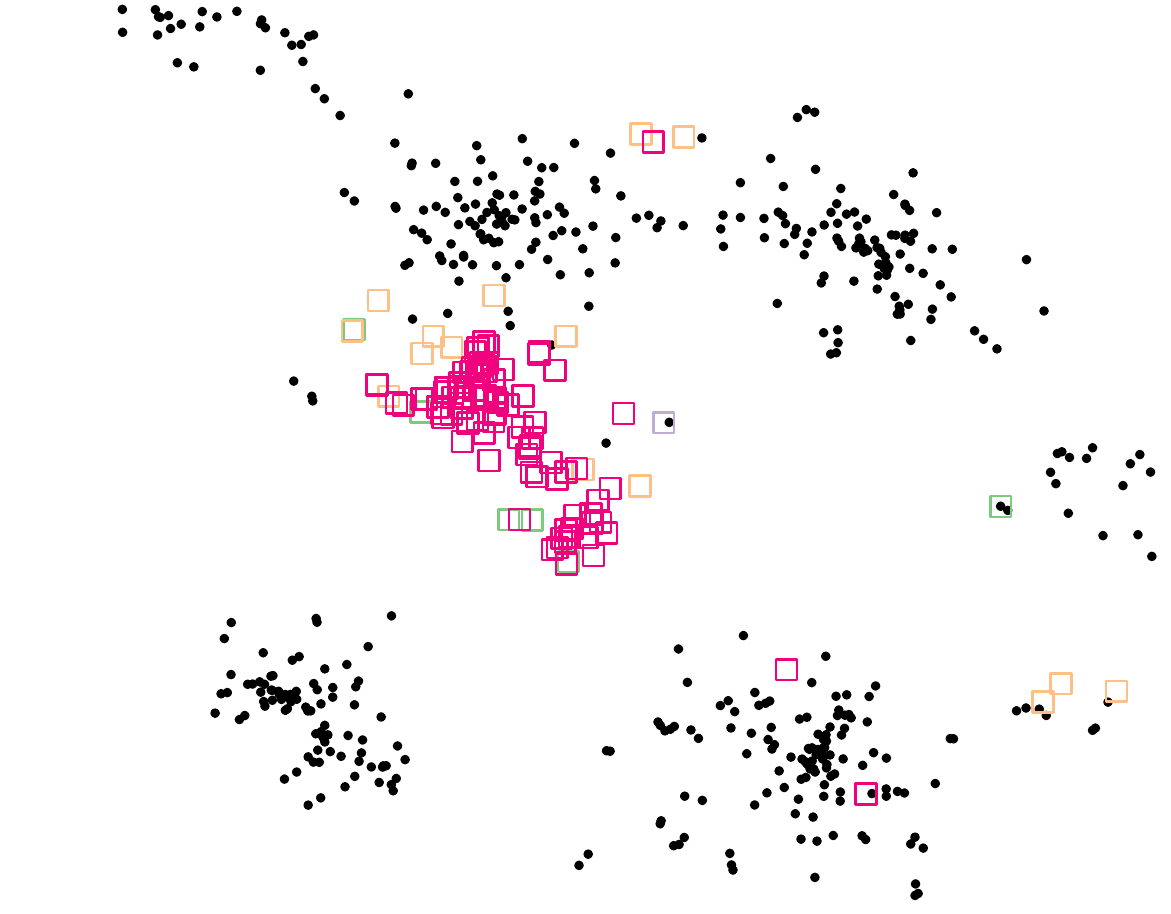}
	}
	\caption{(Best viewed in color) 2D embedding space (a) before and (b) after fine-tuning ResNetS. The black circles indicate base categories of \textit{mini}ImageNet and novel categories are expressed as colored squares. (c)-(g) are enlarged figures to corresponding parts of (b). We observe that the features of novel categories are well placed between base categories.}\label{fig:TSNE}
\end{figure*}

\begin{figure*}
	\centering
	\subfigure
	{
		\centering
		\includegraphics[width=0.16\linewidth]{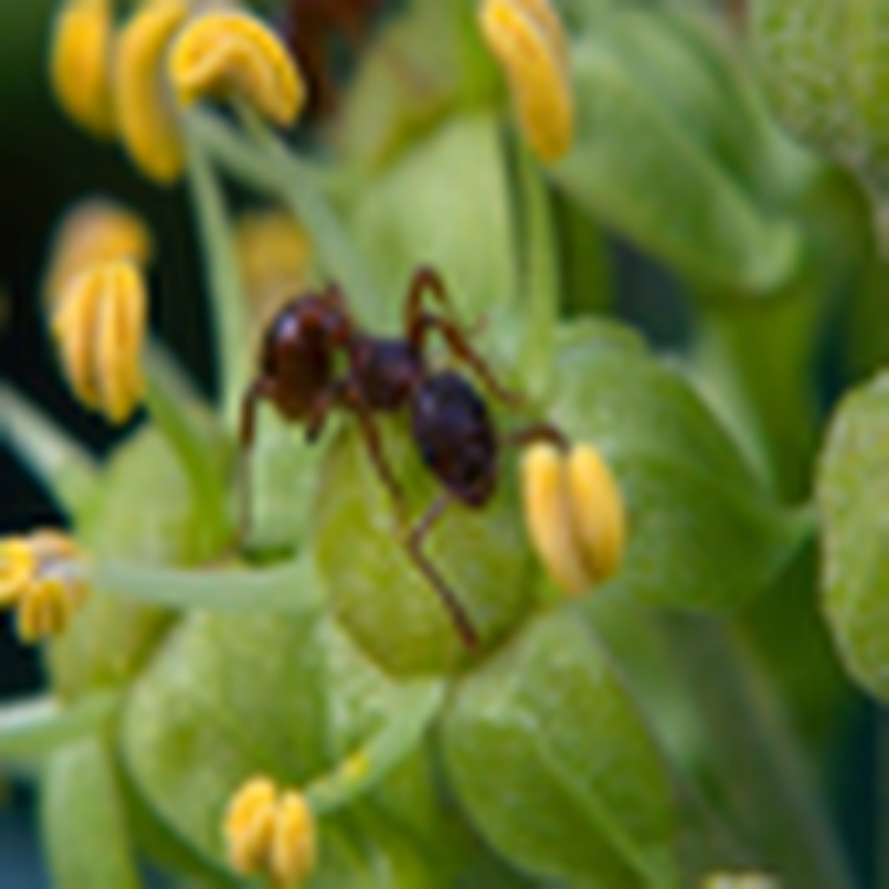}
	} \hspace{-0.47cm}
	\subfigure
	{
		\centering
		\includegraphics[width=0.16\linewidth]{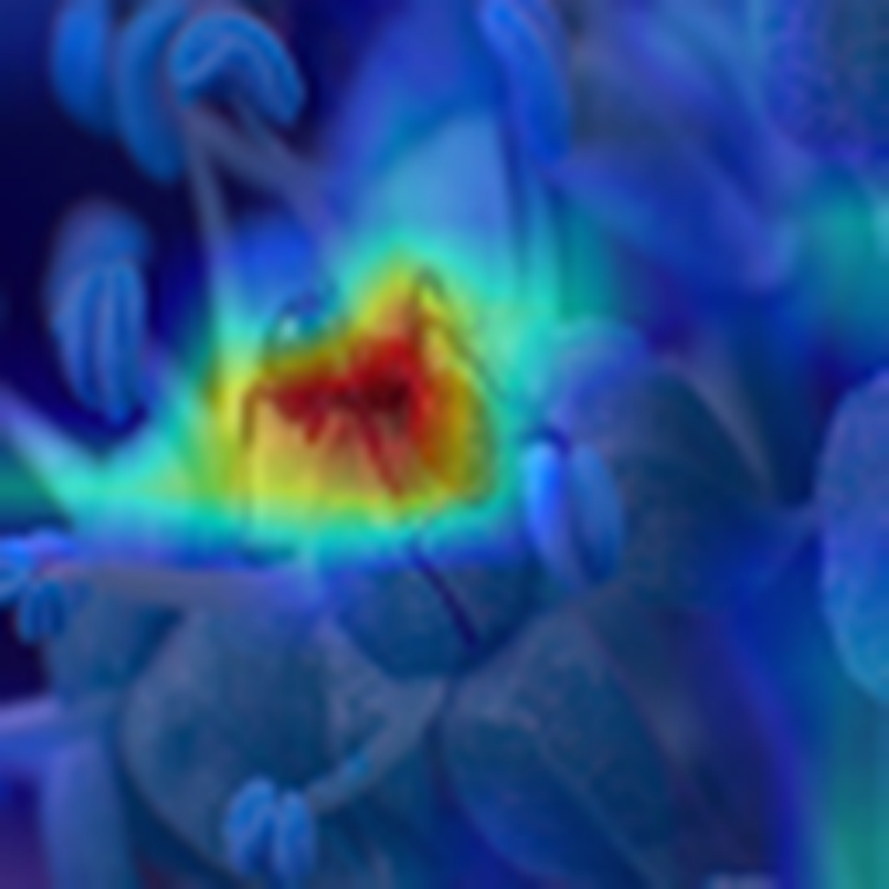}
	} \hspace{-0.47cm}
	\subfigure
	{
		\centering
		\includegraphics[width=0.16\linewidth]{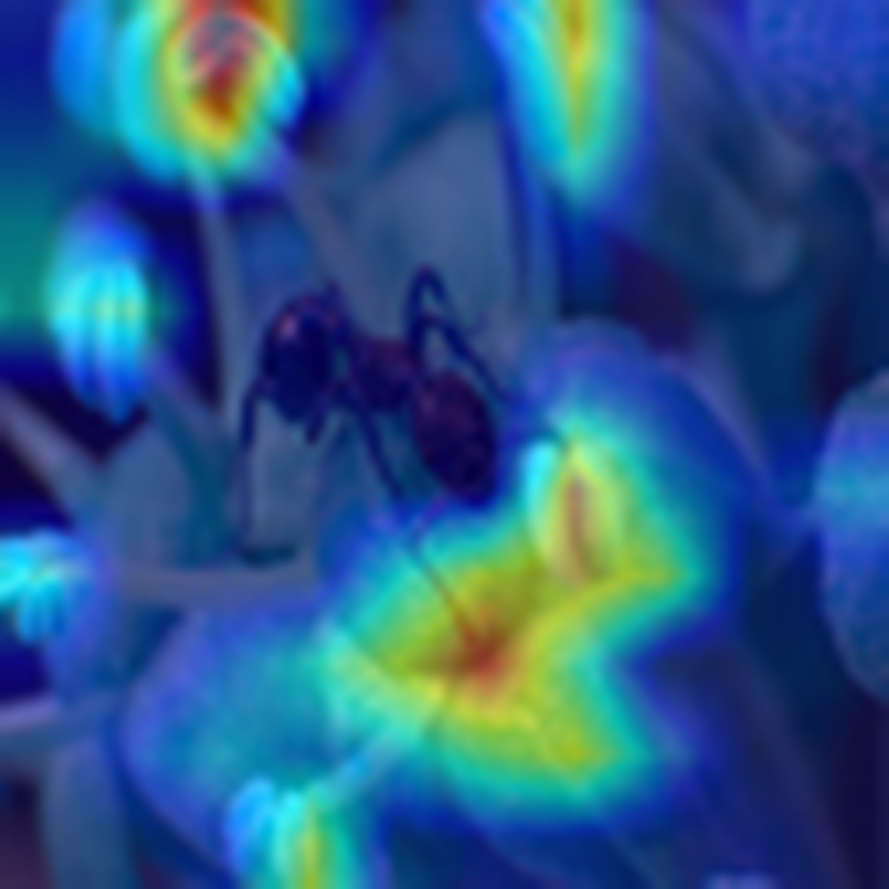}
	} 
	\subfigure
	{
		\centering
		\includegraphics[width=0.16\linewidth]{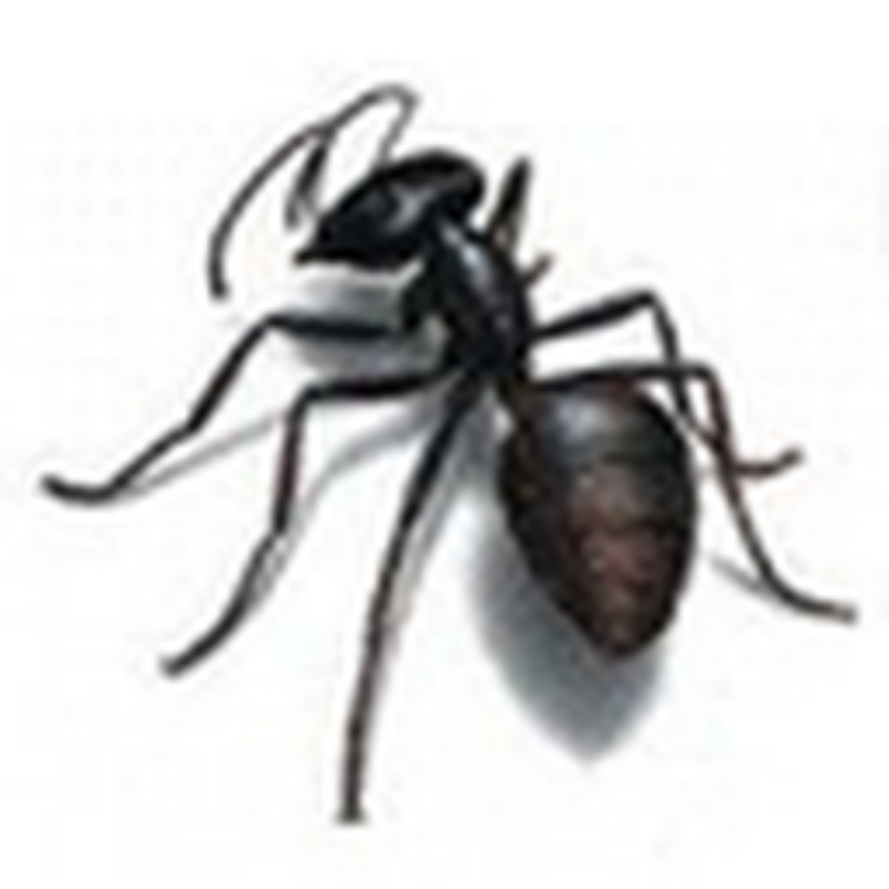}
	} \hspace{-0.47cm}
	\subfigure
	{
		\centering
		\includegraphics[width=0.16\linewidth]{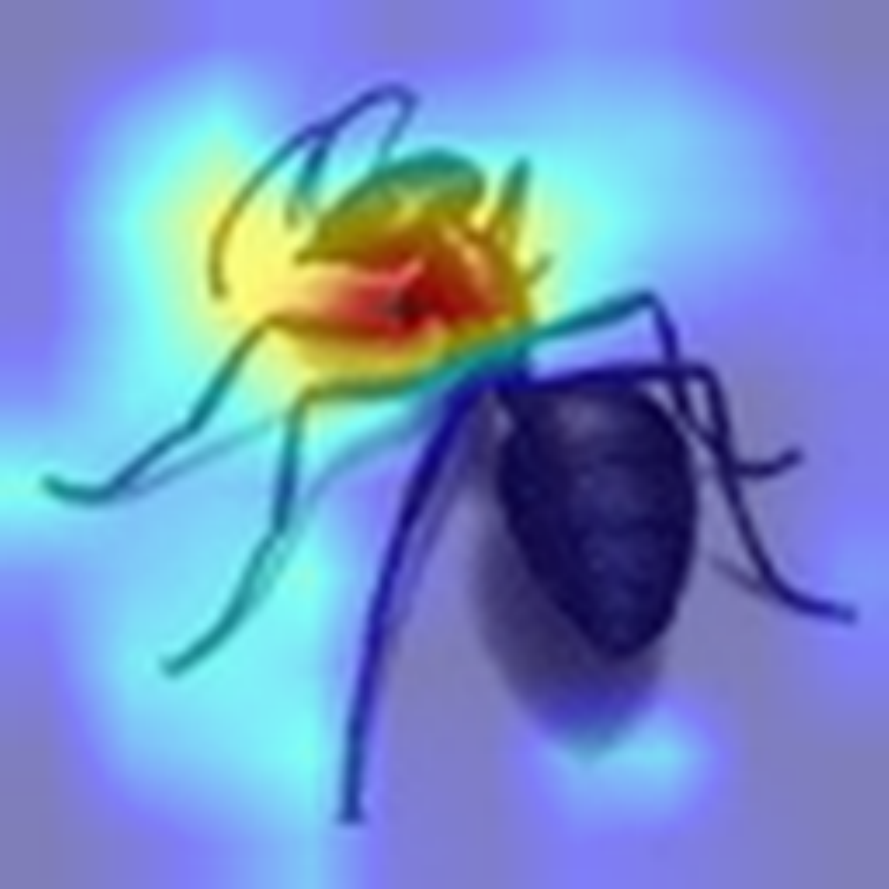}
	} \hspace{-0.47cm}
	\subfigure
	{
		\centering
		\includegraphics[width=0.16\linewidth]{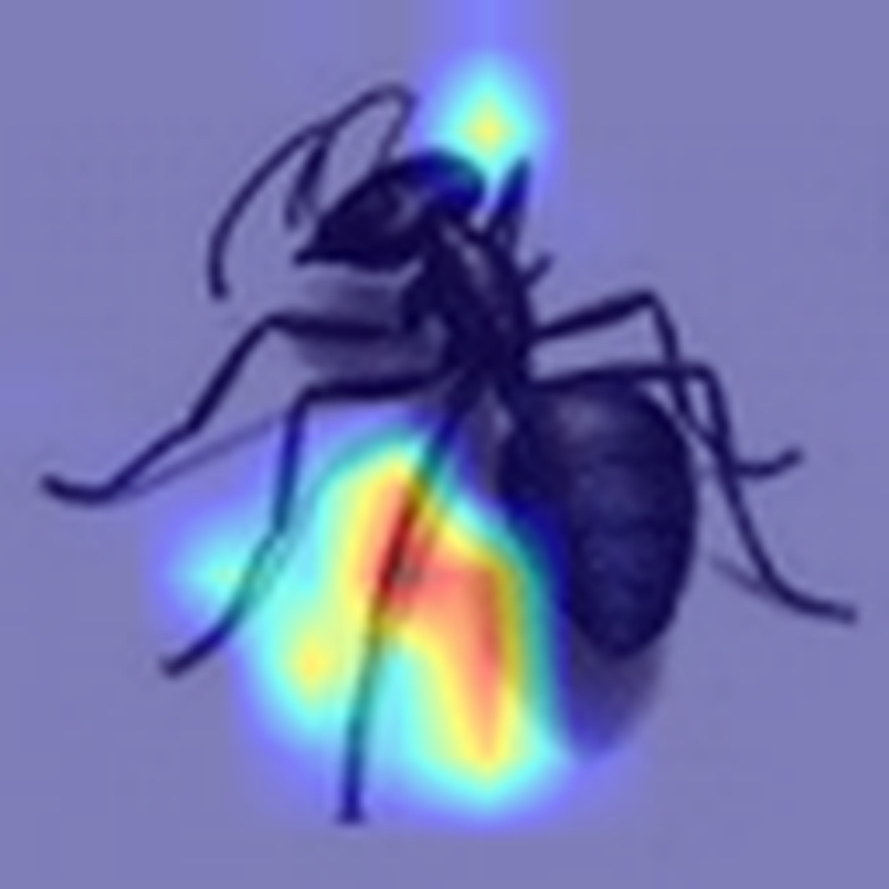}
	}

	\vspace{-0.3cm}

	\subfigure
	{
		\centering
		\includegraphics[width=0.16\linewidth]{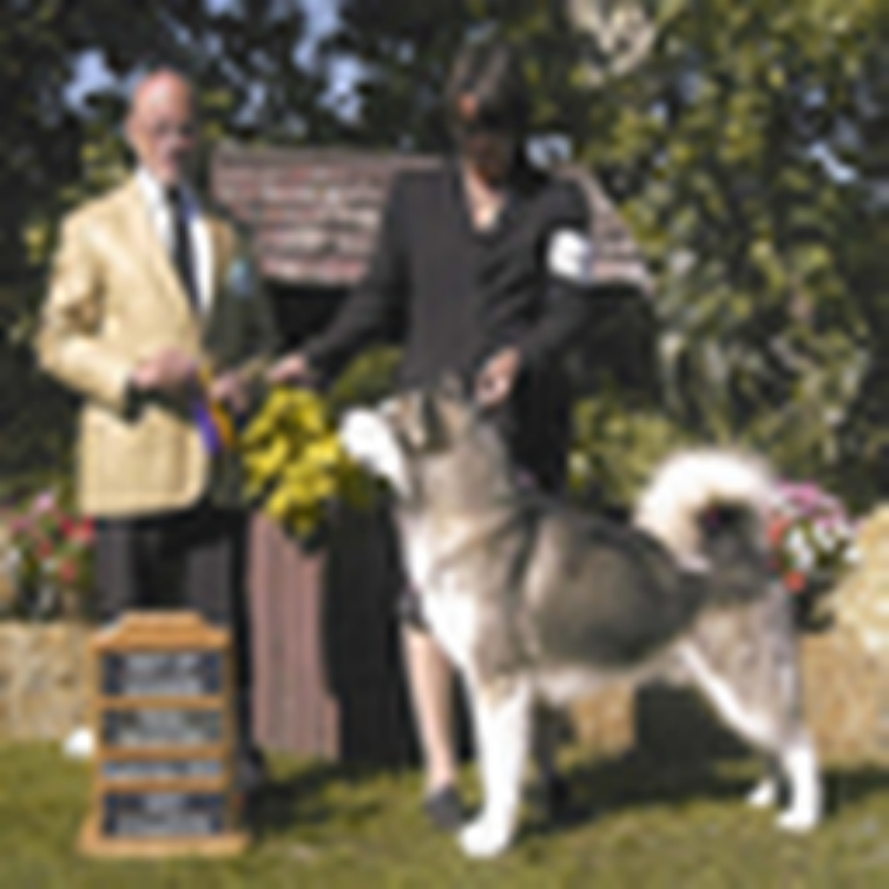}
	} \hspace{-0.47cm}
	\subfigure
	{
		\centering
		\includegraphics[width=0.16\linewidth]{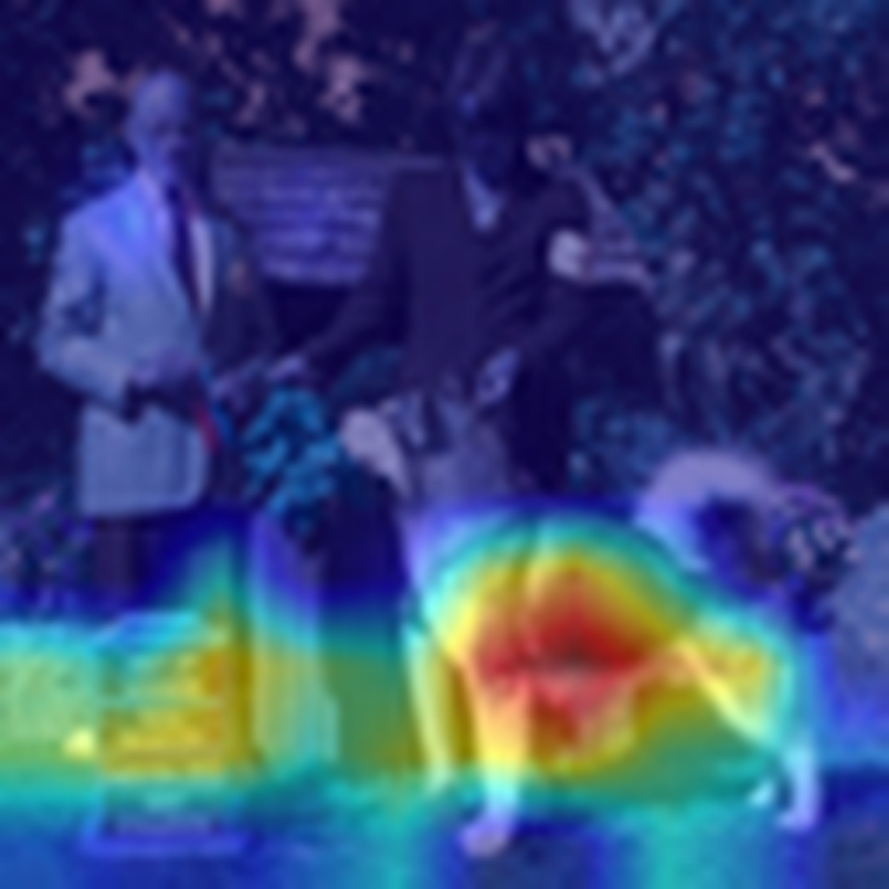}
	} \hspace{-0.47cm}
	\subfigure
	{
		\centering
		\includegraphics[width=0.16\linewidth]{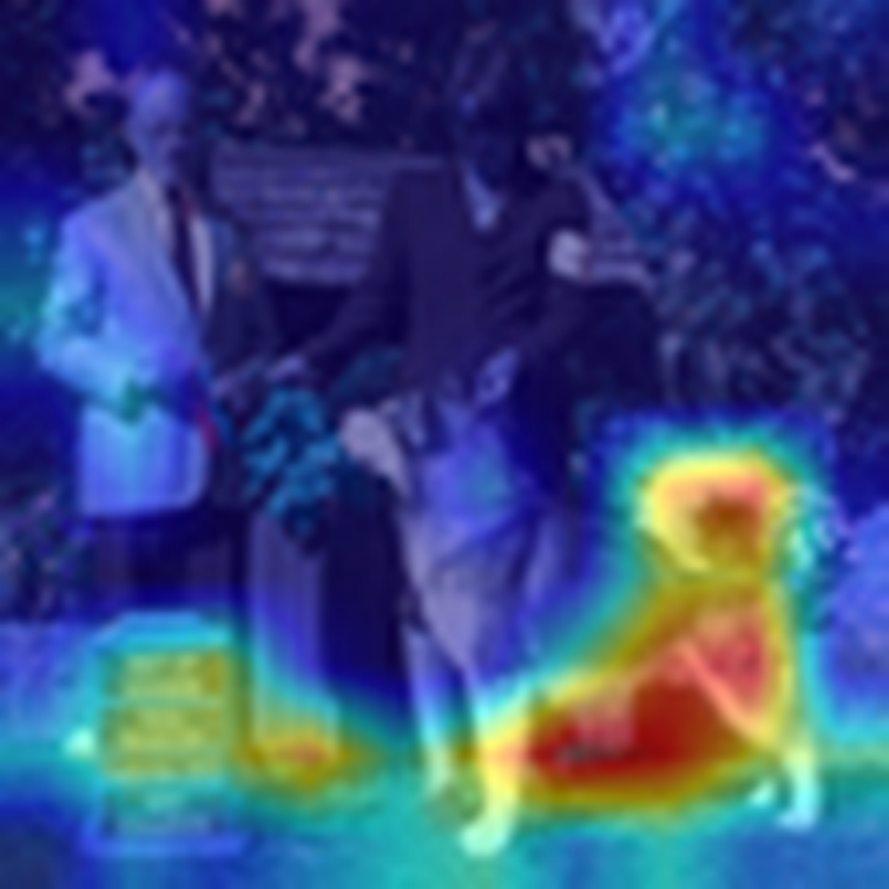}
	}
	\subfigure
	{
		\centering
		\includegraphics[width=0.16\linewidth]{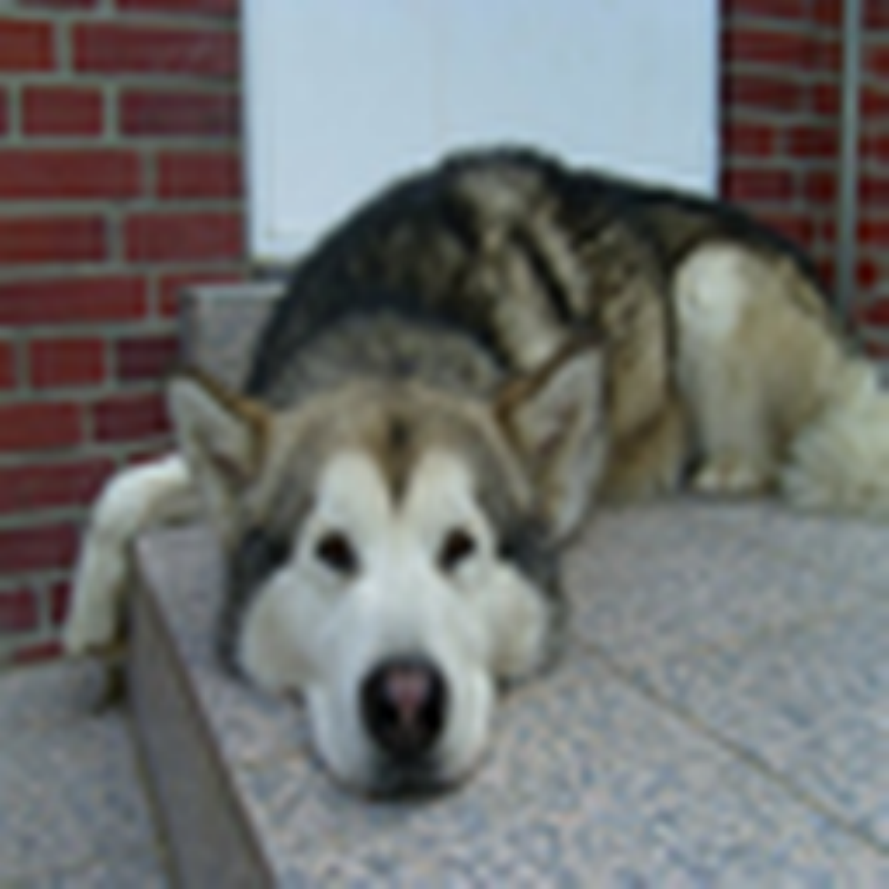}
	} \hspace{-0.47cm}
	\subfigure
	{
		\centering
		\includegraphics[width=0.16\linewidth]{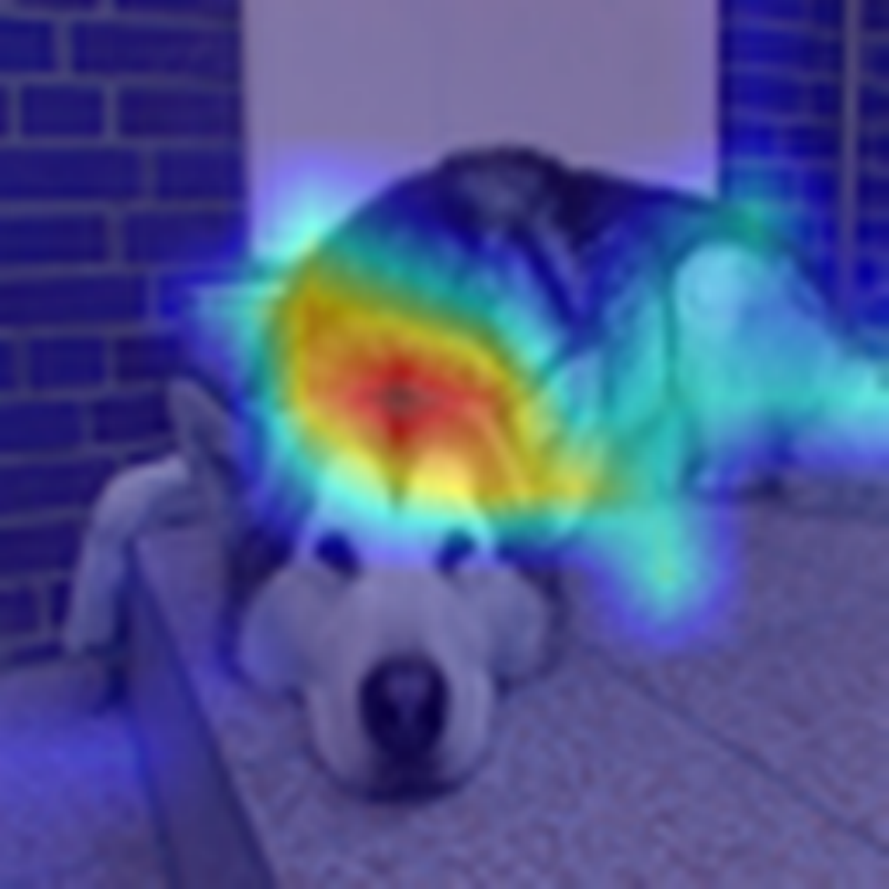}
	} \hspace{-0.47cm}
	\subfigure
	{
		\centering
		\includegraphics[width=0.16\linewidth]{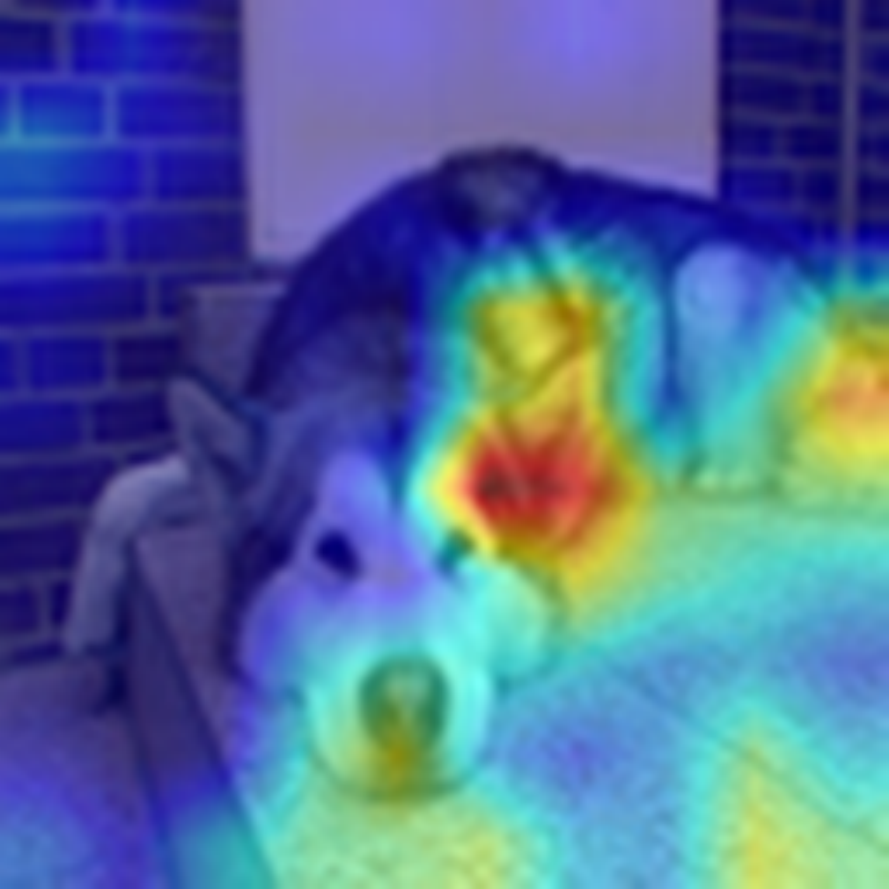}
	}

  	\vspace{-0.3cm}
	\setcounter{subfigure}{0}
	\subfigure[Original]
	{
		\centering
		\includegraphics[width=0.16\linewidth]{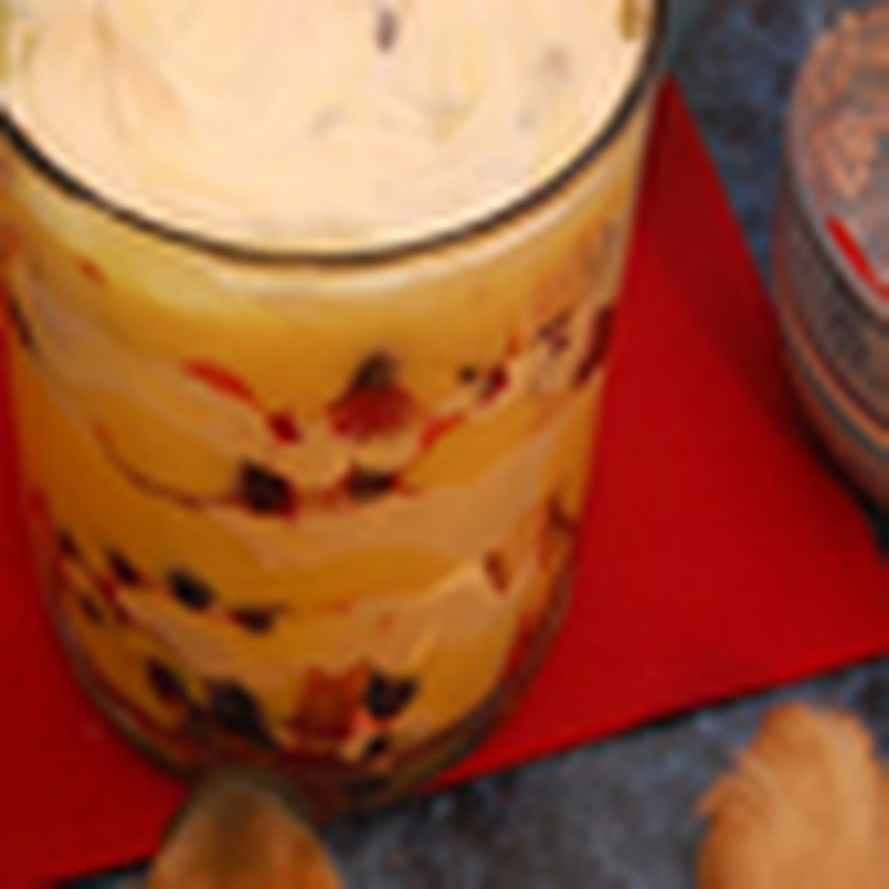}
	} \hspace{-0.47cm}
	\subfigure[After fine-tuning]
	{
		\centering
		\includegraphics[width=0.16\linewidth]{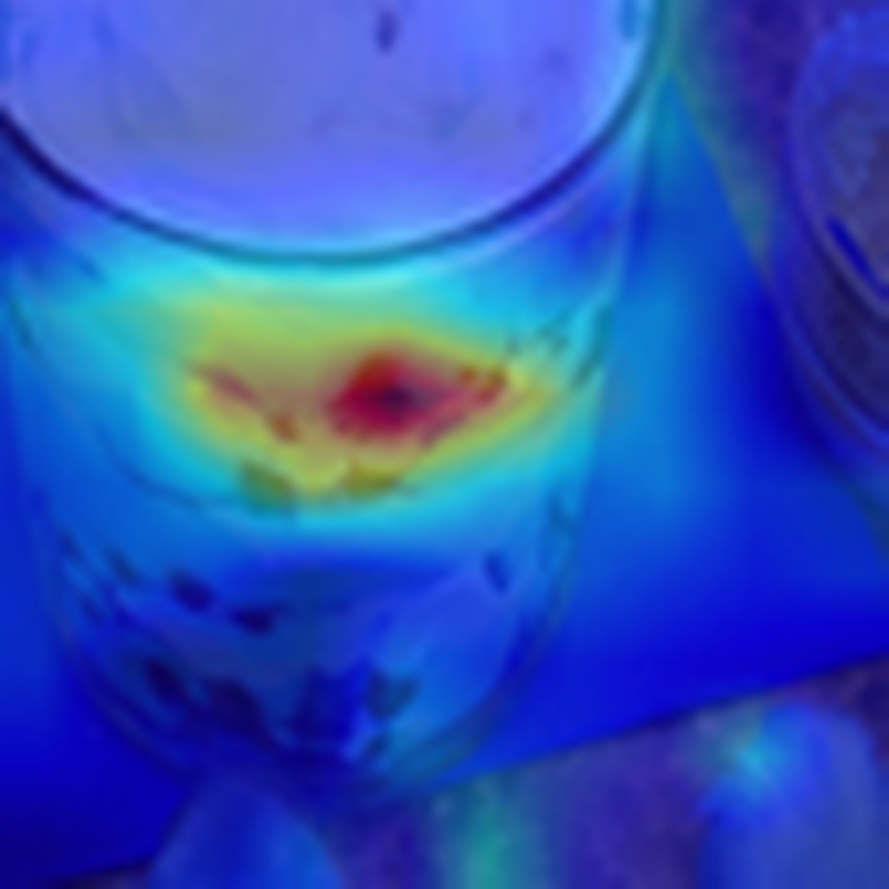}
	} \hspace{-0.47cm}
	\subfigure[Before fine-tuning]
	{
		\centering
		\includegraphics[width=0.16\linewidth]{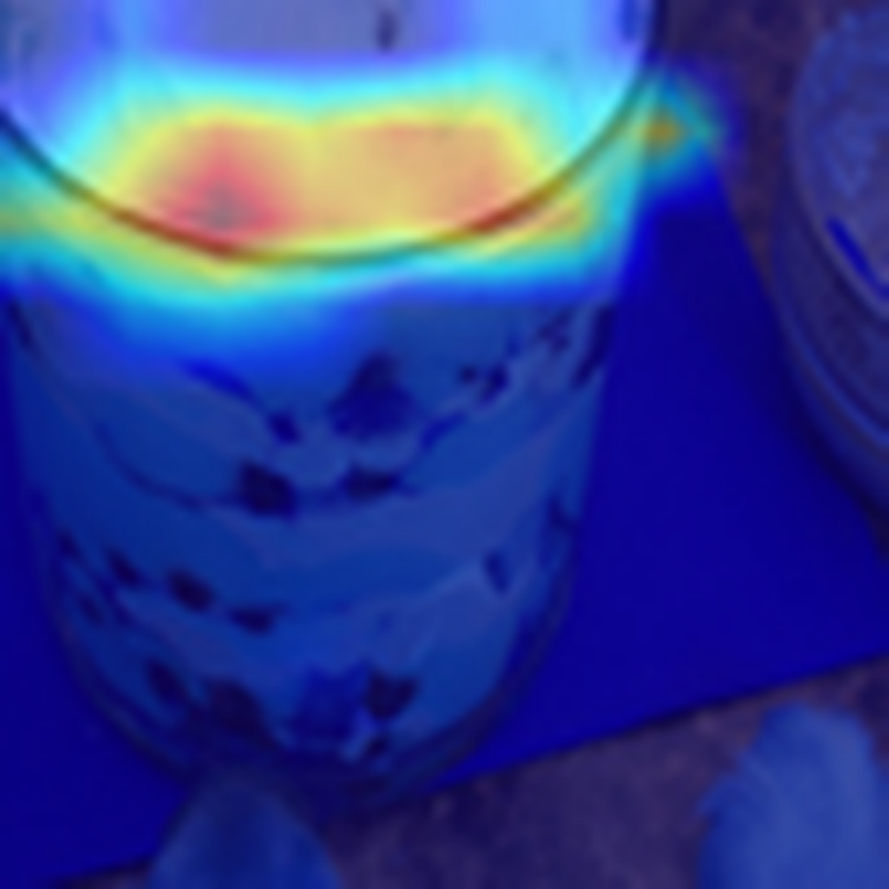}
	}
	\subfigure[Original]
	{
		\centering
		\includegraphics[width=0.16\linewidth]{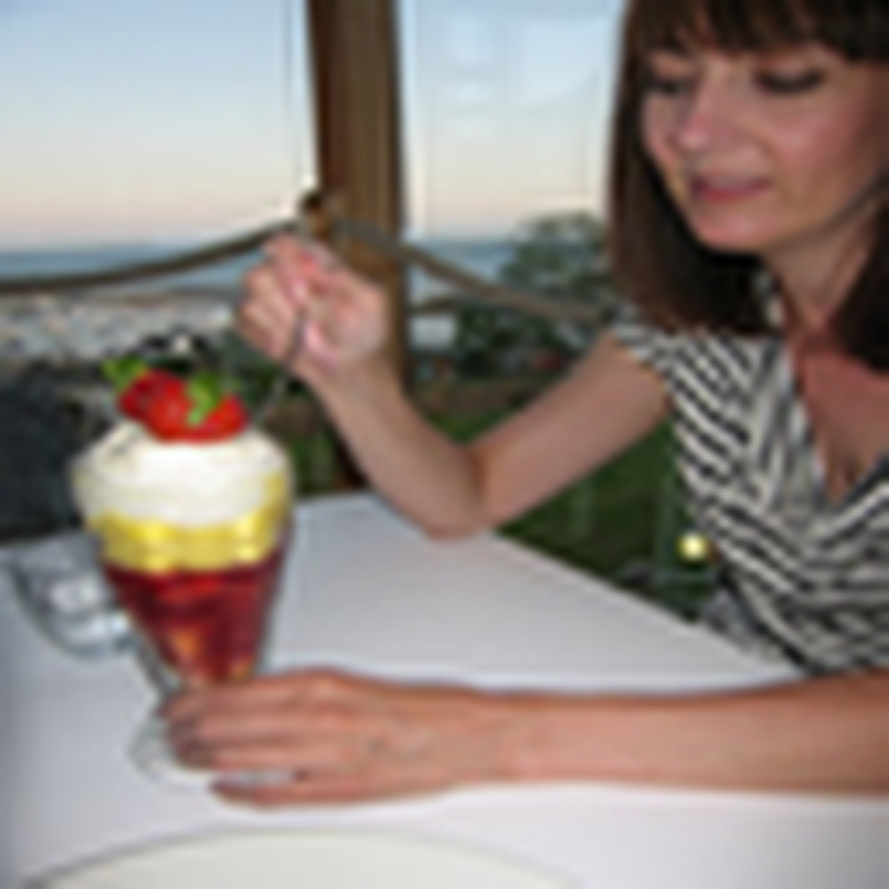}
	} \hspace{-0.47cm}
	\subfigure[After fine-tuning]
	{
		\centering
		\includegraphics[width=0.16\linewidth]{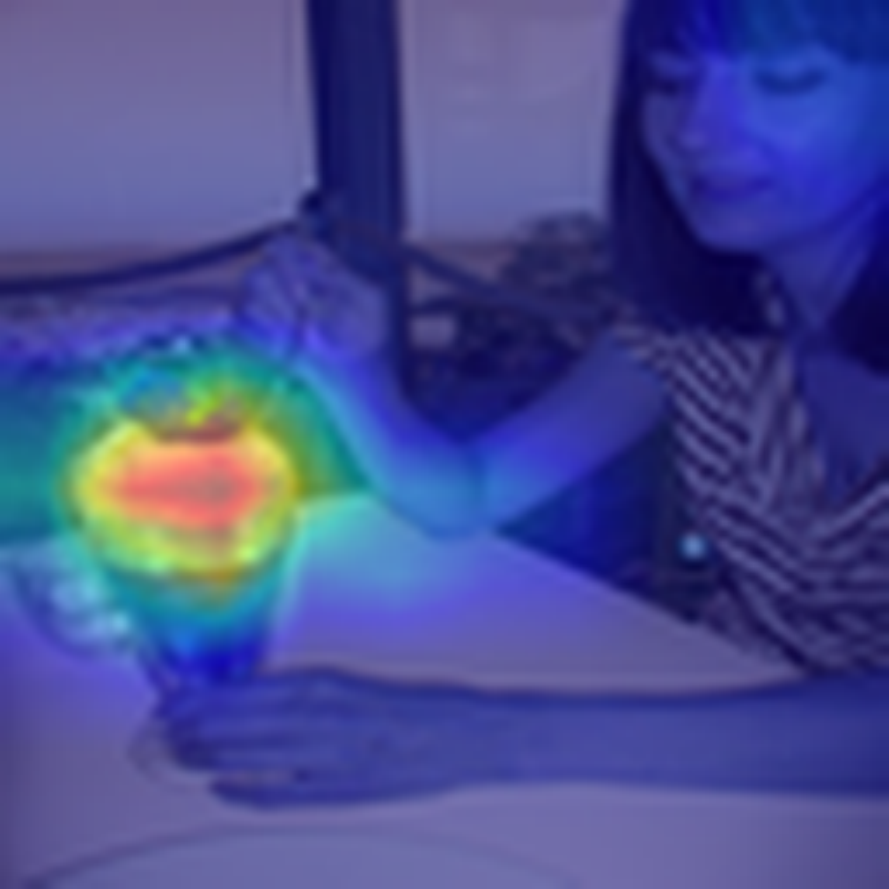}
	} \hspace{-0.47cm}
	\subfigure[Before fine-tuning]
	{
		\centering
		\includegraphics[width=0.16\linewidth]{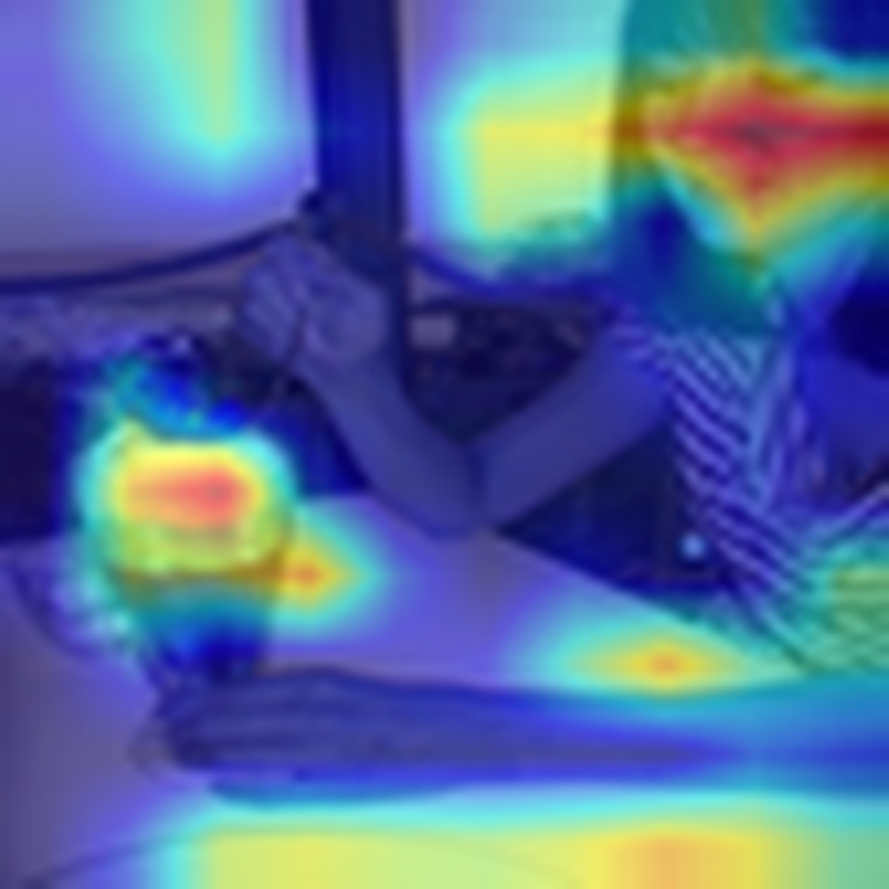}
	}
	\caption{Feature maps for novel categories. (a),(d): original images, (b),(e): feature maps after fine-tuning and (c),(f): feature maps before fine-tuning. Our algorithm focuses on important parts of objects and removes noisy parts.}\label{fig:Feature_map}
\end{figure*}

\begin{figure*}
		\centering
        \subfigure[Cosine similarity of ResNetS]
        {
                \centering
                \includegraphics[width=.30\linewidth, height=5cm]{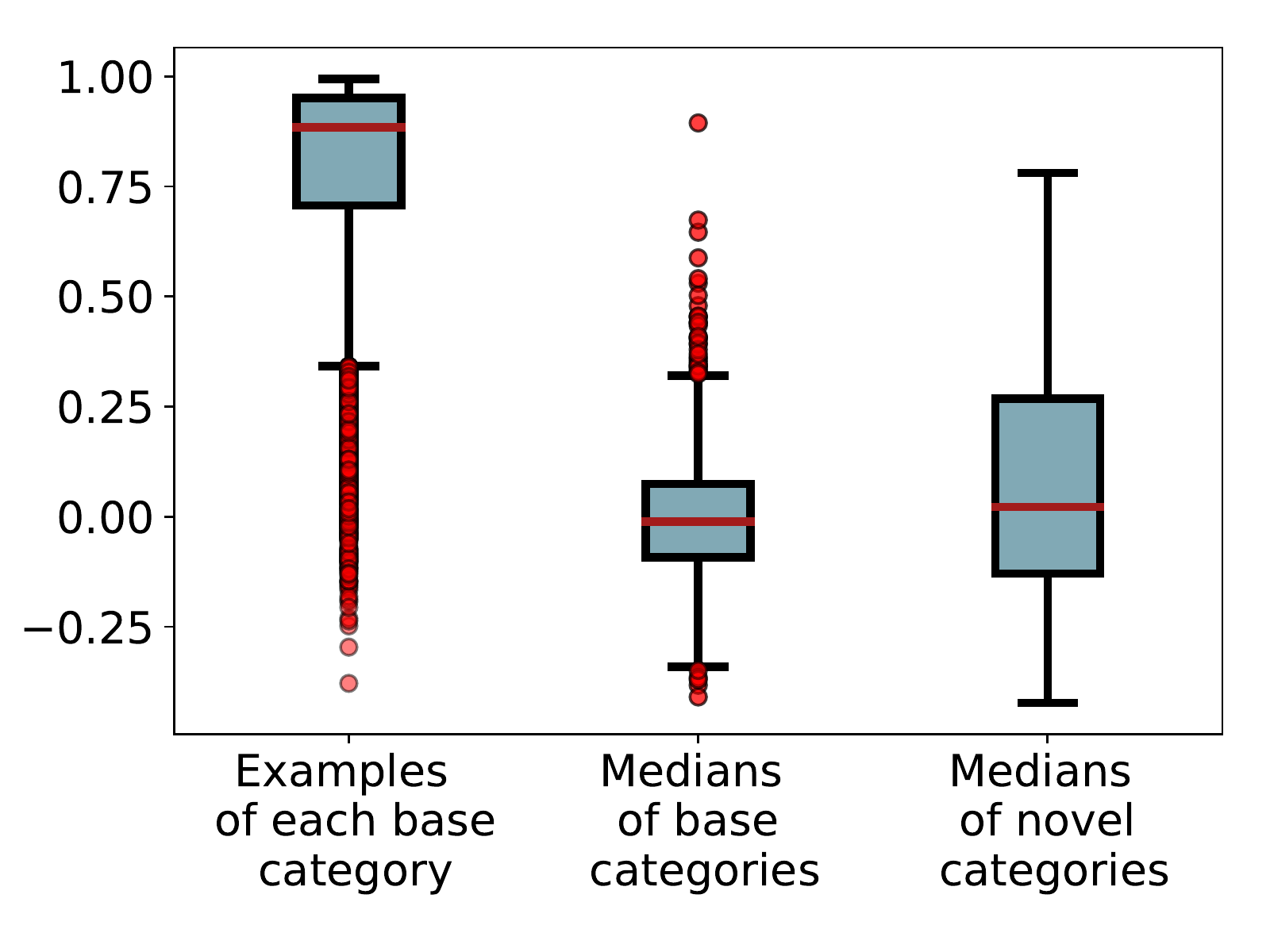}
                \label{fig:ang_resnets}
		}
        \subfigure[Cosine similarity of C64F]
        {
                \centering
                \includegraphics[width=.30\linewidth, height=4.9cm]{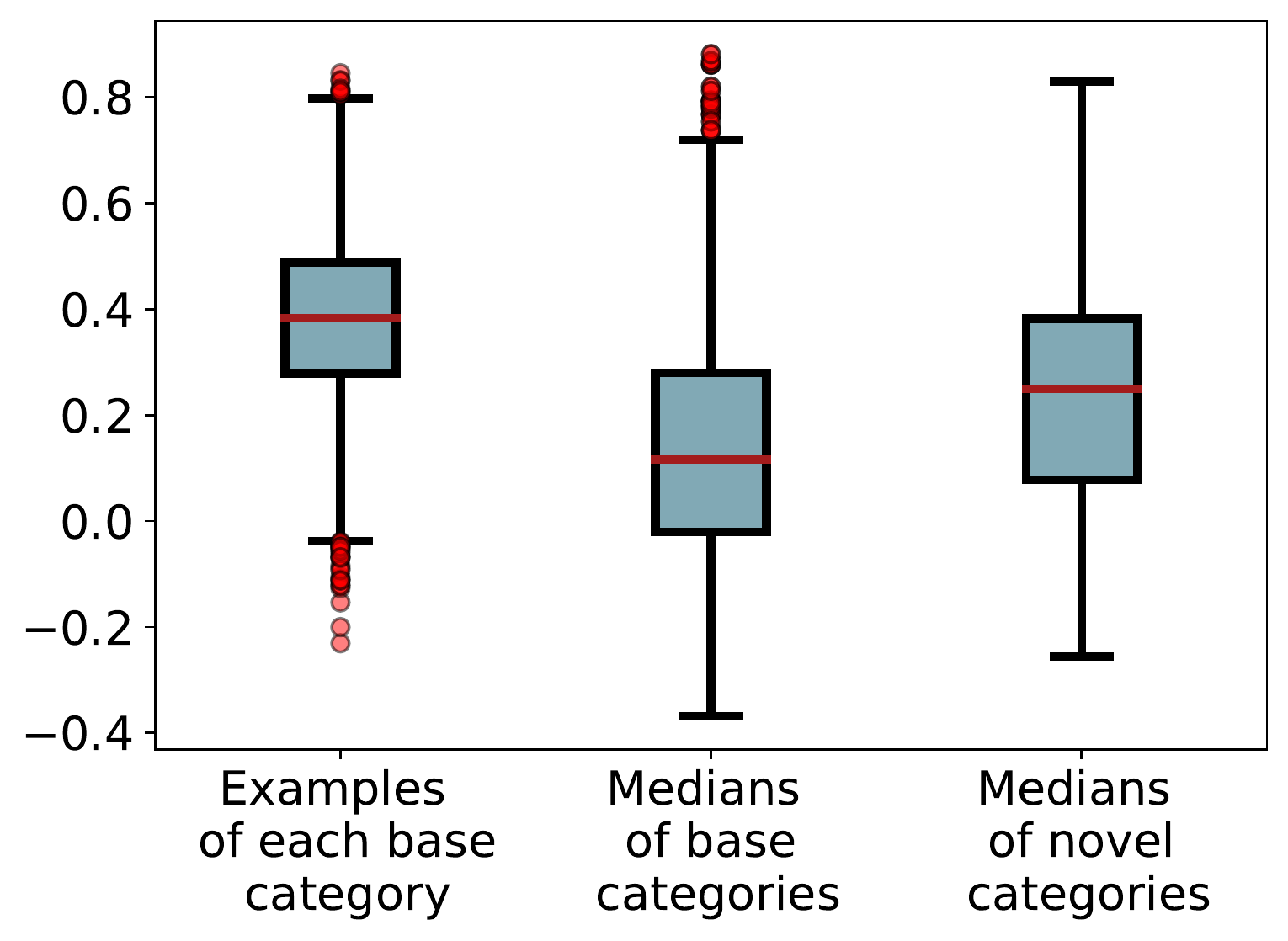}
                \label{fig:ang_c64f}
		}
        \subfigure[Cosine similarity of ResNet10]
        {
                \centering
                \includegraphics[width=.30\linewidth, height=5cm]{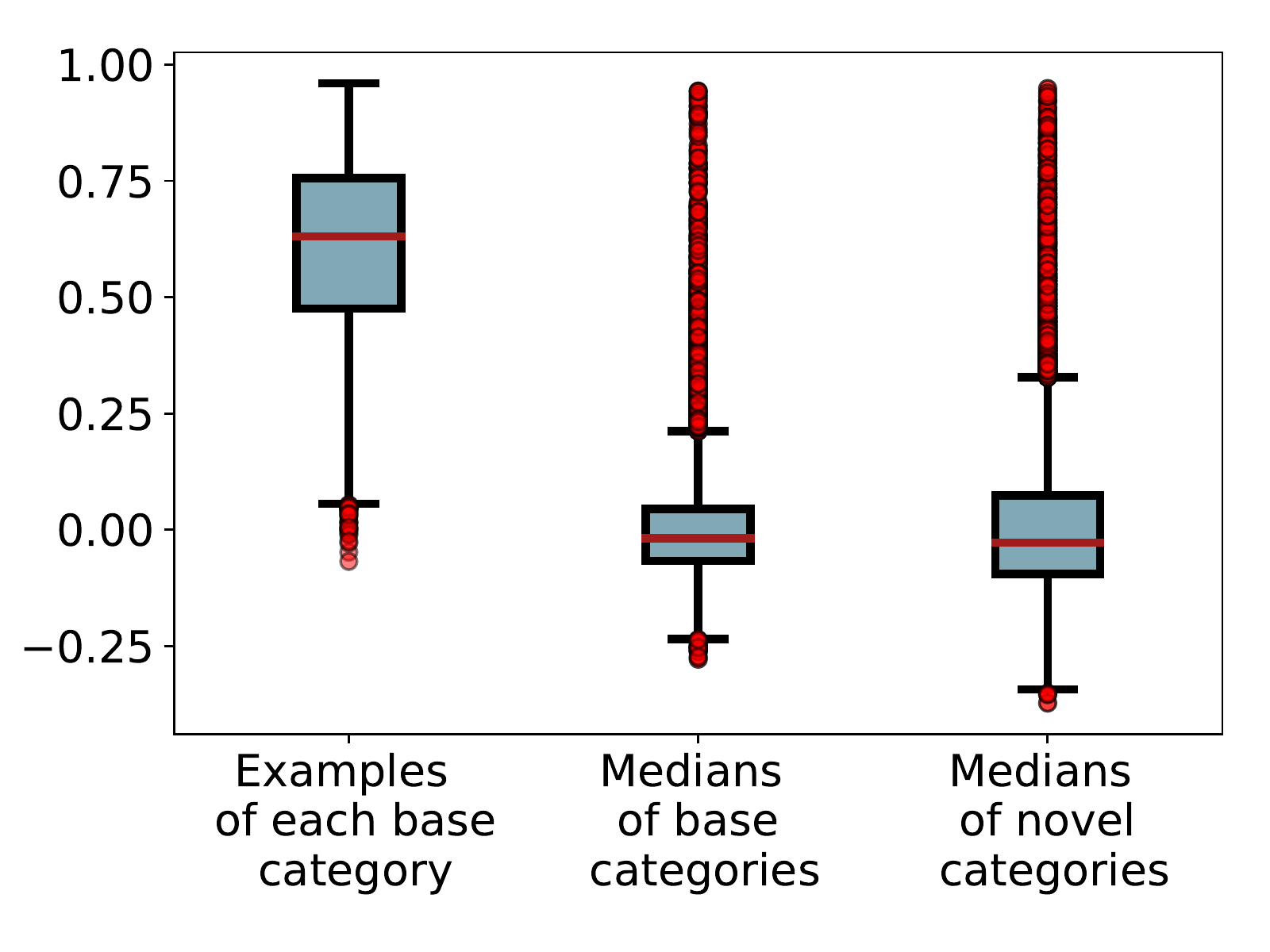}
                \label{fig:ang_resnet10}
		}
        \caption{Box plots of cosine similarity. Given a network trained on base categories, we first extracted features of the examples for base and novel categories. Based on the feature space, we obtained the medians of the features of each category. For figures (a)-(c), we calculated the cosine similarity among (left) examples belonging to each base category, (middle) medians of base categories and (right) medians of novel categories.}\label{fig:ang_for_each_network}
\end{figure*}

\setlength\dashlinedash{1pt}
\setlength\dashlinegap{1.5pt}
\setlength\arrayrulewidth{0.3pt}
\begin{table*}
\caption{Comparison between the proposed method and the single-stage fine-tuning approach on the ImageNet with ResNet10-Dropout. For base categories, the average accuracy of all shots is reported.}\label{table:soft_large}
\centering
\resizebox{\textwidth}{!}{
 \renewcommand{\arraystretch}{1.2}
    \large
    \begin{tabular}{c|ccccc|cccccc|cccccc|cc}
    \hline
    \multirow{2}[2]{*}{Models} & \multicolumn{5}{c|}{Novel}                                                                        &                   & \multicolumn{5}{c|}{Both}                                                                         &                   & \multicolumn{5}{c|}{Both with prior}                                                              &                   & \multirow{2}[2]{*}{Base} \bigstrut[t]\\
                      & k=1               & 2                 & 5                 & 10                & 20                &                   & k=1               & 2                 & 5                 & 10                & 20                &                   & k=1               & 2                 & 5                 & 10                & 20                &                   &  \bigstrut[b]\\
    \hline
    \multirow{2}[2]{*}{Proposed-Dropout} & 49.57             & 60.89             & 71.07             & 76.11             & 78.84             &                   & 60.39             & 67.44             & 74.22             & 77.32             & 79.05             &                   & 58.91             & 65.84             & 72.71             & 75.96             & 77.88             &                   & \multirow{2}[2]{*}{\textbf{93.55}} \bigstrut[t]\\
                      & $\pm$.24              & $\pm$.16              & $\pm$.09              & $\pm$.07              & $\pm$.05              &                   & $\pm$.14              & $\pm$.10              & $\pm$.06              & $\pm$.05              & $\pm$.03              &                   & $\pm$.14              & $\pm$.10              & $\pm$.06              & $\pm$.05              & $\pm$.04              &                   &  \bigstrut[b]\\
    \hline
    Single-stage      & 45.08             & 57.01             & 69.98             & 75.1              & 78.04             &                   & 55.23             & 63.34             & 71.08             & 73.82             & 75.24             &                   & 54.44             & 61.55             & 69.76             & 73.27             & 75.34             &                   & \multirow{2}[2]{*}{\textbf{90.45}} \bigstrut[t]\\
    fine-tuning       & $\pm$.43              & $\pm$.28              & $\pm$.16              & $\pm$.13              & $\pm$.09              &                   & $\pm$.34              & $\pm$.20              & $\pm$.13              & $\pm$.08              & $\pm$.05              &                   & $\pm$.29              & $\pm$.20               & $\pm$.14              & $\pm$.07              & $\pm$.05              &                   &  \bigstrut[b]\\
    \hline
    \end{tabular}%
}
\end{table*}

\subsection{Graphical Analysis}
We aim to extract discriminative features for the novel categories while preserving the feature space learned for the base categories. To show that our algorithm accords with the purpose, we visualize the feature space before and after applying our fine-tuning method on \textit{mini}ImageNet with ResNetS. We used T-SNE \cite{maaten2008visualizing} for dimensionality reduction. As shown in Fig. \ref{fig:TSNE}, the features of novel categories are well placed between base categories. Without fine-tuning, features of a category are spread out. This spreadability implies that it is crucial to fine-tune a feature extractor for some novel categories. 

Furthermore, we visualize feature maps for novel categories using Grad-CAM \cite{selvaraju2017grad} that is a visualization method to analyse what deep networks learn. As shown in Fig. \ref{fig:Feature_map}, our algorithm focuses on important parts of objects and removes noisy parts, which helps improve the classification accuracy.

\section{Discussion}\label{sec:discussion}
We first discuss why the proposed method significantly outperforms the state of the art on \textit{mini}ImageNet and is slightly better on the ImageNet. Then, we discuss the following questions: Does a single-stage fine-tuning approach work? How does the performance vary according to the possible combinations of the geometric constraints in Eq. \ref{eqn:tot}? How sensitive is the performance to the margin value in Eq. \ref{eqn:AWS}? Finally, we discuss the limitation of the proposed method.

\setlength\dashlinedash{1pt}
\setlength\dashlinegap{1.5pt}
\setlength\arrayrulewidth{0.3pt}
\begin{table}
\caption{Comparison between the proposed method and the single-stage fine-tuning approach on \textit{mini}ImageNet with ResNetS.}\label{table:soft_mini}
\centering
\resizebox{\columnwidth}{!}{
 \renewcommand{\arraystretch}{1.2}
    \large
    \begin{tabular}{c|ccc|ccc}
    \hline
    \multirow{2}[2]{*}{Models} & \multicolumn{3}{c|}{5-way 5-shot}                         & \multicolumn{3}{c}{5-way 1-shot} \bigstrut[t]\\
                      & Novel             & Both              & Base              & Novel             & Both              & Base \bigstrut[b]\\
    \hline
    Proposed          & 78.00$\pm$0.61        & 68.16             & \textbf{79.78}    & 58.52$\pm$0.82        & 56.05             & \textbf{79.78} \bigstrut\\
    \hline
    Single-stage      & \multirow{2}[2]{*}{76.78$\pm$0.59} & \multirow{2}[2]{*}{58.62} & \multirow{2}[2]{*}{\textbf{64.58}} & \multirow{2}[2]{*}{55.11$\pm$0.78} & \multirow{2}[2]{*}{52.95} & \multirow{2}[2]{*}{\textbf{72.32}} \bigstrut[t]\\
    fine-tuning       &                   &                   &                   &                   &                   &  \bigstrut[b]\\
    \hline
    \end{tabular}%
}
\end{table}

\subsection{Network Characteristic}
Since performance gains over other methods vary depending on the networks, we analyse the reason in terms of network characteristics. Given a network trained on base categories, we first extracted features of the examples for base and novel categories. Based on the feature space, we obtained the medians of the features of each category. For example, if we have $64$ categories, this process produces $64$ medians corresponding to each category. Then, we calculated the cosine similarity among (a) features of the examples belonging to each base category, (b) the medians of base categories and (c) the medians of novel categories. Specifically, (a) is to show how well the features cluster for a category they belong to. (b) and (c) are to investigate how sufficiently the categories are separated from each other. For statistical analysis, we used box plots. As shown in Fig. \ref{fig:ang_for_each_network}, features from ResNetS well cluster together for each category and categories are separated sufficiently. However, a lot of categories are close together on C64F and ResNet10. 

Based on the above characteristics, we compare the proposed method with the novel weight generator by Gidaris and Komodakis  \cite{gidaris2018dynamic} that shows comparable results to ours. To train the novel weight generator \cite{gidaris2018dynamic}, the algorithm samples some base categories and regards them as novel categories. This is to mimic the environment when real novel categories are given. For this training procedure to be effective and generalized, the distribution of features for base categories should resemble that of novel categories. However, this is not the case with ResNetS. As a result, despite the much higher capacity, the performance of ResNetS is worse than that of C64F. On the contrary, our proposed method relies on the feature discriminability trained on base categories. Thus, the proposed method produces valuable results with ResNetS but is less effective with C64F and ResNet10. Even though performance gains over other methods vary depending on network characteristic, we achieve the best performance on all datasets.

\subsection{Single-Stage Fine-Tuning Approach}
To validate that a simple fine-tuning approach with only a few examples destroys a pre-trained network, we present the following experimental setting: after training a network using base categories, we added novel weights to the classifier. Given a few training examples, we then fine-tuned the novel weights of the classifier and the last convolution block of the feature extractor with cross-entropy loss. For this experiment, the feature extractor was shared for base and novel categories. We used SGD with learning rates $1e$-4 for ResNetS and $1e$-5 for ResNet10. As shown in Tables \ref{table:soft_mini} and \ref{table:soft_large}, the single-stage approach destroys the pre-trained network. Especially, the performance on base categories cannot be preserved unlike the proposed method.

\subsection{Effectiveness of each Loss Function}
\begin{figure*}
		\centering
        \subfigure[Both categories with ResNetS on \textit{mini}ImageNet]
        {
                \centering
                \includegraphics[width=.48\linewidth]{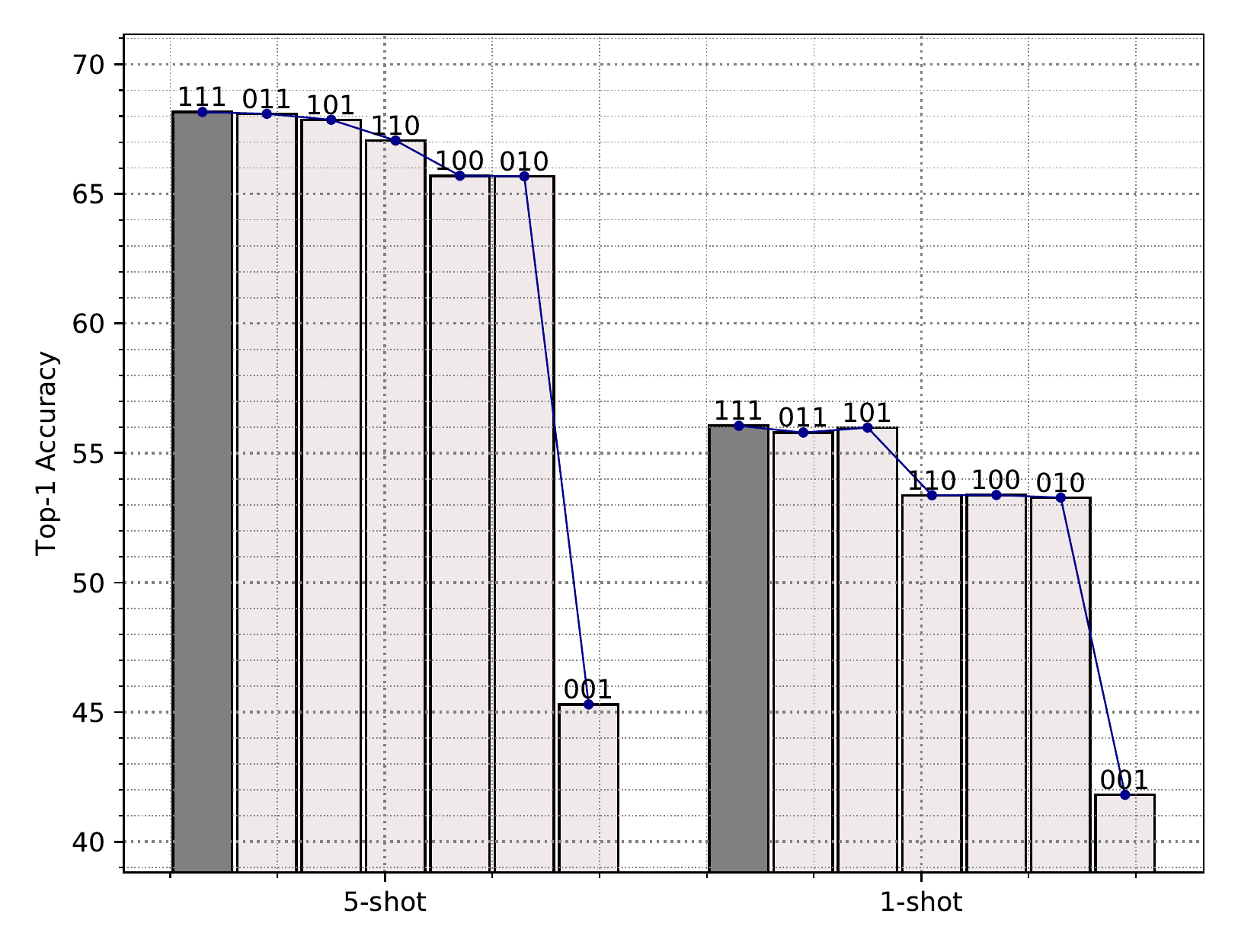}
                \label{fig:reg_resnets_both}
		}
        \subfigure[Novel categories with ResNetS on \textit{mini}ImageNet]
        {
                \centering
                \includegraphics[width=.48\linewidth]{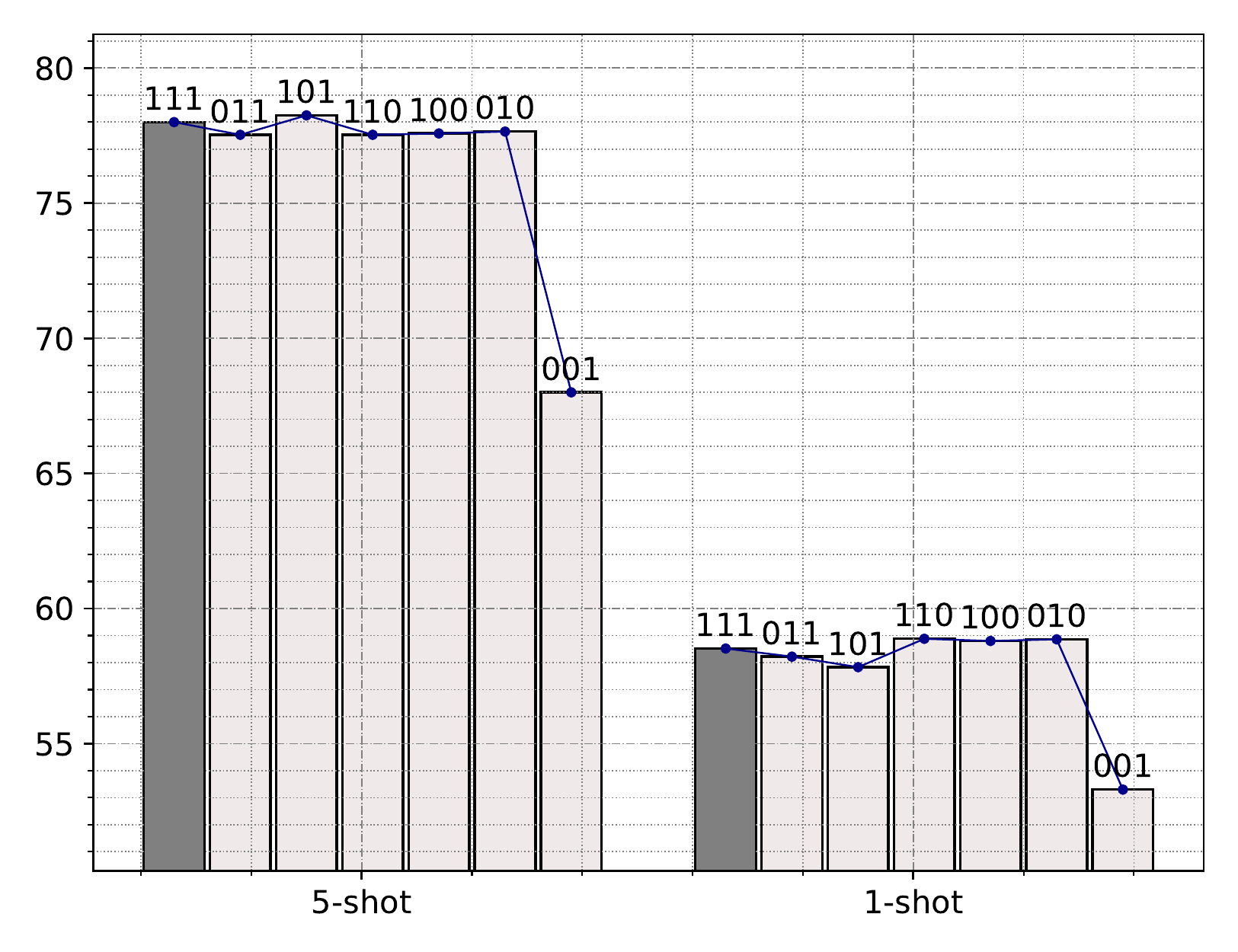}
                \label{fig:reg_resnets_novel}
		}
        \subfigure[Both categories with ResNet10-Dropout on the ImageNet]
        {
                \centering
                \includegraphics[width=.96\linewidth]{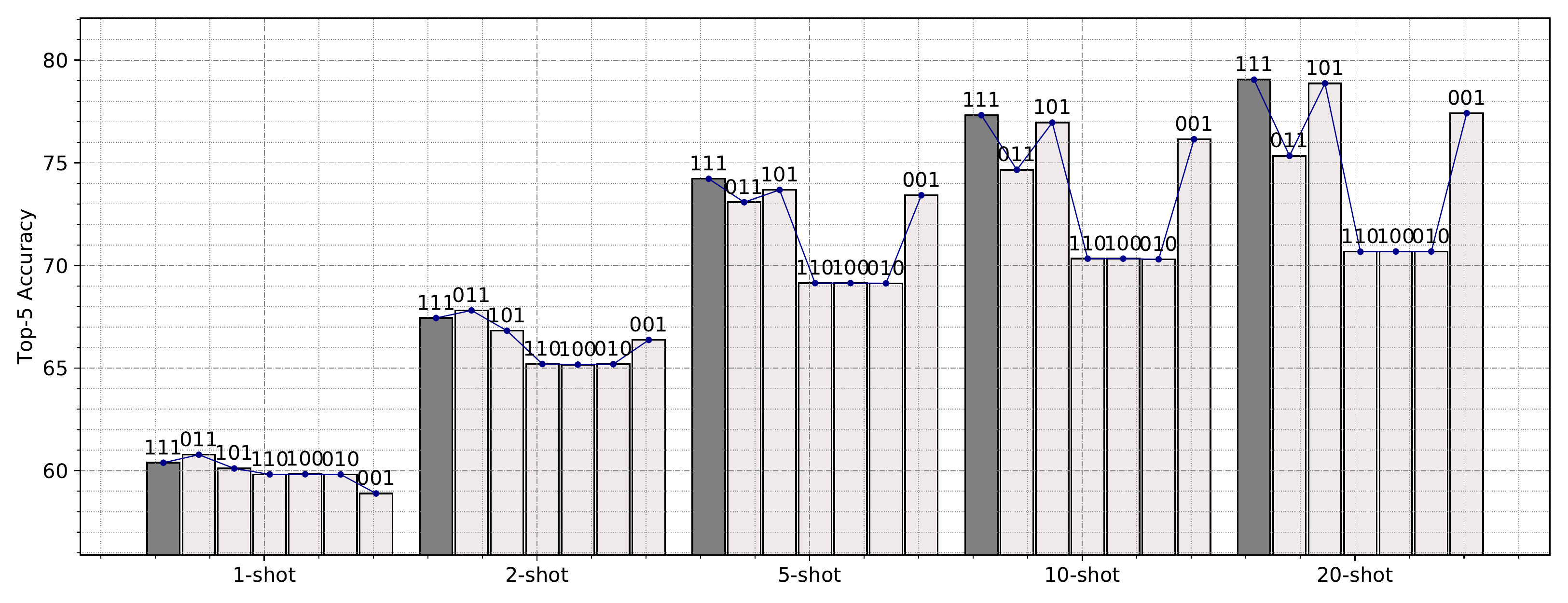}
                \label{fig:reg_resnets10_both}
		}
        \subfigure[Novel categories with ResNet10-Dropout on the ImageNet]
        {
                \centering
                \includegraphics[width=.96\linewidth]{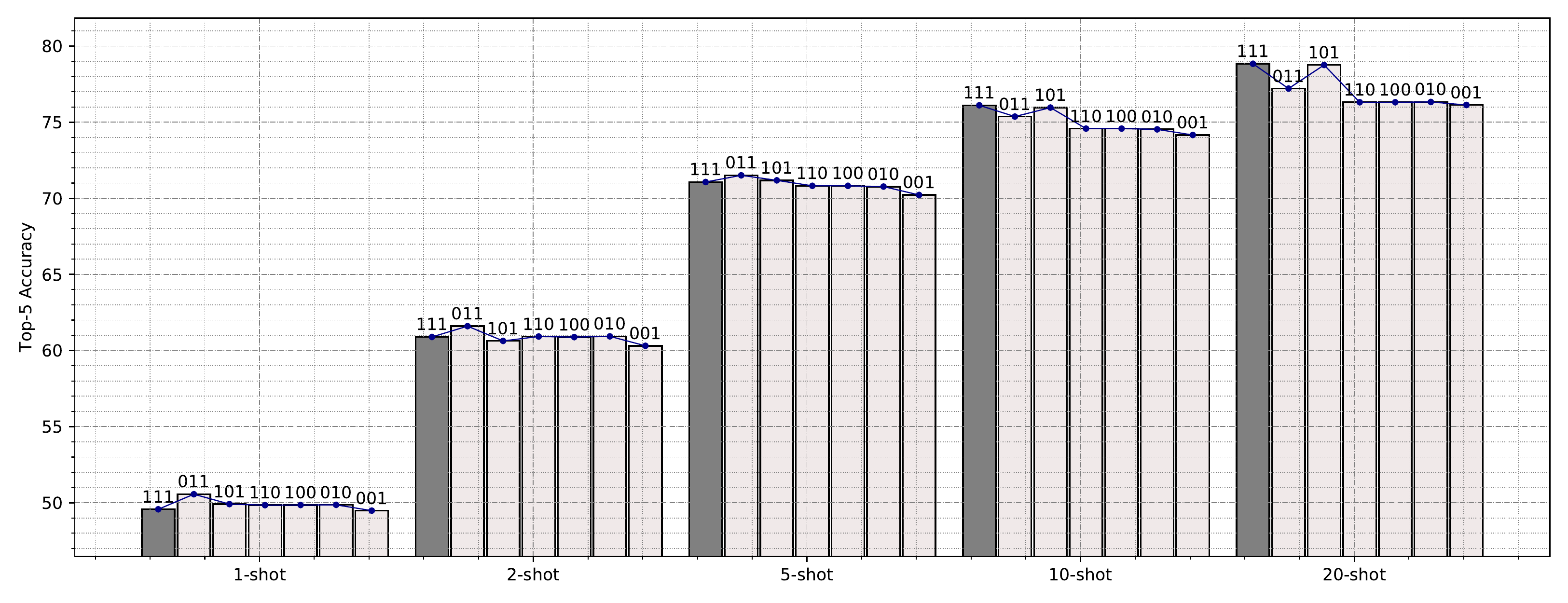}
                \label{fig:reg_resnet10_novel}
		}				
        \caption{Classification accuracy according to the combinations of the three loss functions. The binary numbers correspond to $\gamma\alpha\beta$ in Eq. \ref{eqn:tot}. The gray bar indicates $111$, which means we use cross entropy loss, the weight-centric feature clustering and the angular weight separation.}\label{fig:reg_resnets}
\end{figure*}

Since we use the three loss functions in Eq. \ref{eqn:tot}, the effect of each loss function can be discussed. The accuracy for all possible combinations of the loss functions is shown in Fig. \ref{fig:reg_resnets}. We can observe the following trends. (a) For both categories, the angular weight separation plays a crucial role in the performance. (b) For novel categories, the use of all loss functions does not always result in the best performance. However, it stably produce comparable results for all experiments and the importance increases as the number of training examples increases.

\subsection{Sensitivity of the Margin Value}
\begin{figure*}
		\centering
        \subfigure[Both categories with ResNetS on \textit{mini}ImageNet]
        {
                \centering
                \includegraphics[width=.48\linewidth]{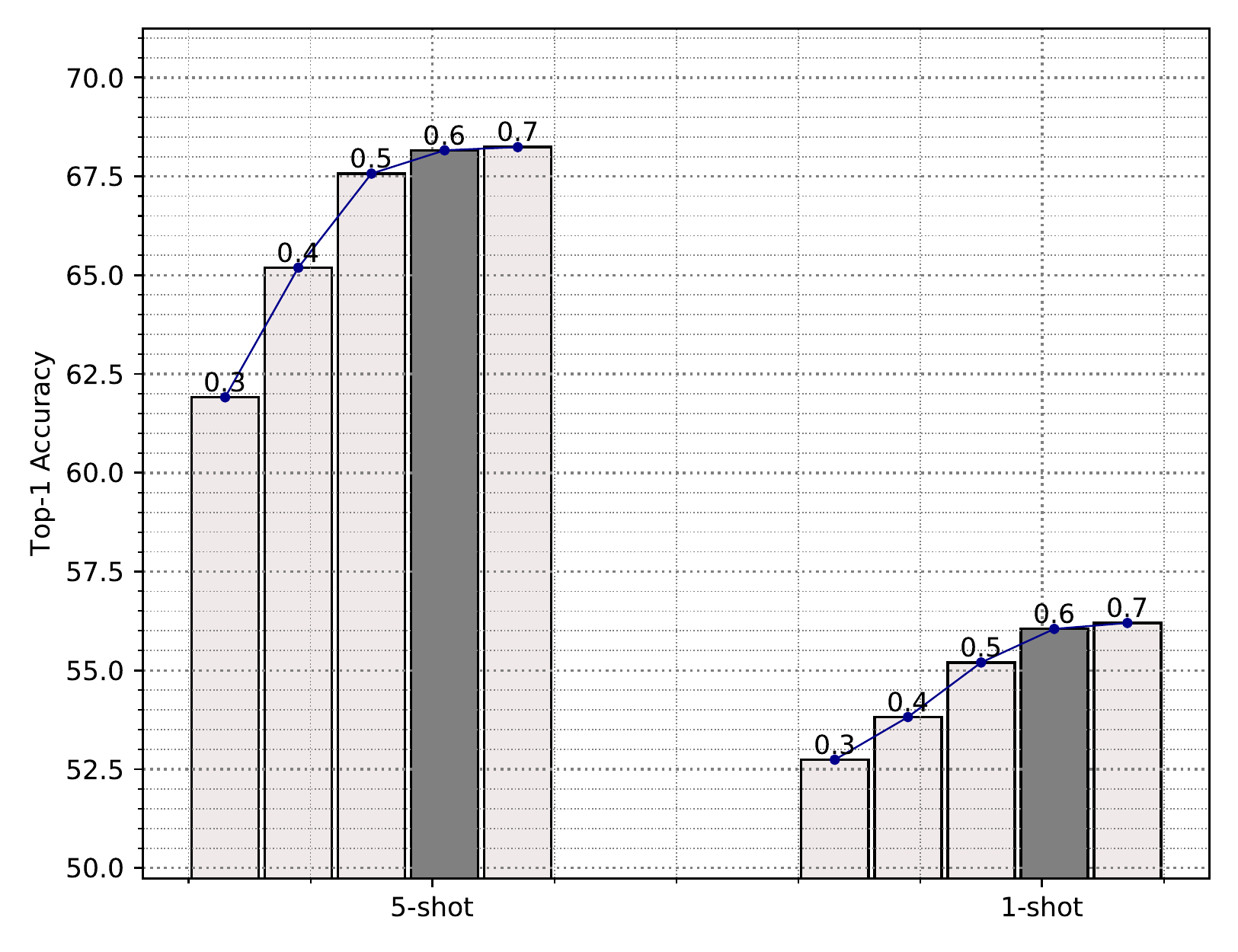}
                \label{fig:margin_resnets_both}
		}
        \subfigure[Novel categories with ResNetS on \textit{mini}ImageNet]
        {
                \centering
                \includegraphics[width=.48\linewidth]{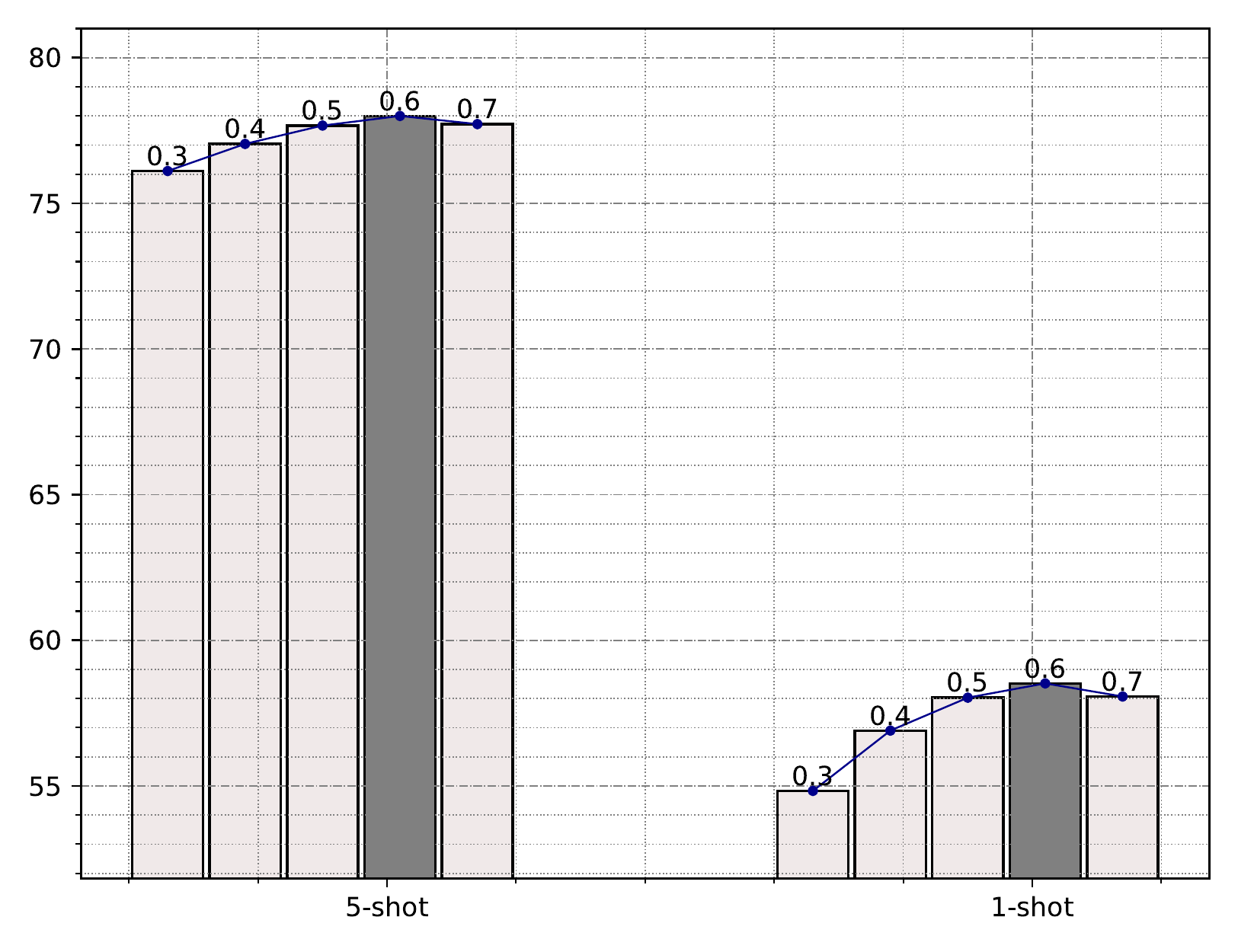}
                \label{fig:margin_resnets_novel}
		}
        \subfigure[Both categories with ResNet10-Dropout on the ImageNet]
        {
                \centering
                \includegraphics[width=.96\linewidth]{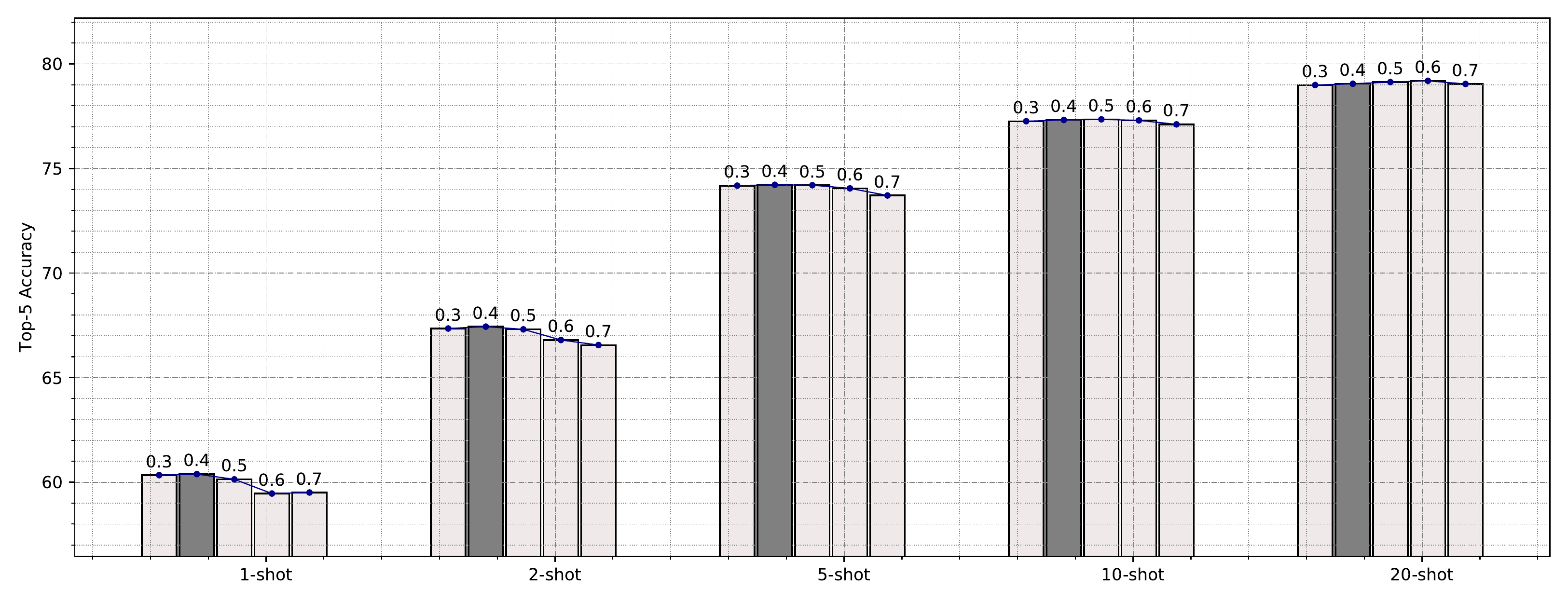}
                \label{fig:margin_resnets10_both}
		}
        \subfigure[Novel categories with ResNet10-Dropout on the ImageNet]
        {
                \centering
                \includegraphics[width=.96\linewidth]{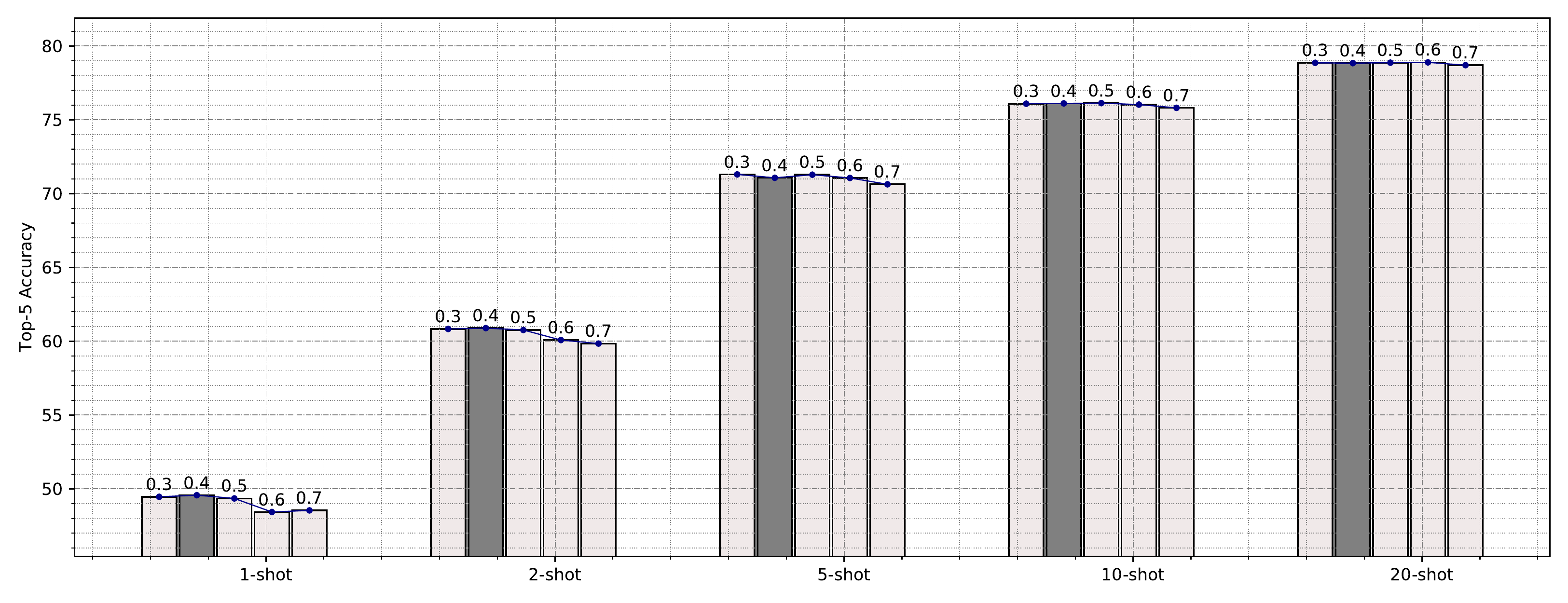}
                \label{fig:margin_resnet10_novel}
		}				
        \caption{Classification accuracy according to various margin values. This margin value is used for the angular weight separation in Eq. \ref{eqn:AWS}. The gray bar indicates the accuracy we reported in Tables \ref{table:mini} and \ref{table:large}.}\label{fig:margin_resnets}
\end{figure*}

For the experiments, we have shown the results with the margin value that was empirically set for the angular weight separation. The purpose of the margin is to maintain the classification weights far from each other. Thus, the value should be determined according to the geometry of the feature space. Nonetheless, Fig. \ref{fig:margin_resnets} shows that the performance is not too sensitive to the variation of the margin value. In other words, the performance does not vary irregularly and this enable us to find a peak point through validation examples.

\subsection{Limitation}
Although the proposed method achieves the state-of-the art performance, it relies on the network capacity. Since we have only a few training examples, it is difficult to adjust features for both base and novel categories. This is why we fine-tune only the parameters for novel categories and freeze the parameters for base categories. In this sense, to boost the performance, we need a network well trained on base categories where the features belonging to a category cluster together and features for different categories are sufficiently separated. To overcome the limitation, although inefficient, training examples for base categories can be used together to adjust the features of both base and novel categories. However, we have only a few training examples for novel categories as ever, which causes imbalanced categories. Thus, we believe that in any case we need a data generation technique for few-shot learning. We would like to leave this for the future work.

\section{Conclusion and future work}\label{sec:conclusion}
In this paper, we presented a fine-tuning strategy for few-shot learning. We considered a network trained for base categories with a large number of training examples and we aimed to add novel categories to it that had only a few training examples. We proposed two geometric constraints to extract discriminative features for the novel categories while preserving the feature space learned for the base categories. The first constraint enabled features of the novel categories to cluster near the category weights when combined with cross-entropy loss. The second maintained the weights of the novel categories far from the weights of the base categories. Applying the constraints, we showed that the accuracy on base categories was not affected even after fine-tuning the network and comparable performances to the base categories could be achieved on novel categories. The proposed method gained state-of-the art performance especially using a network with Dropout. Since we verified a simple way to augment training examples improves the performance, future work involves developing example generation techniques that are advantageous to our idea.

\setlength\dashlinedash{1pt}
\setlength\dashlinegap{1.5pt}
\setlength\arrayrulewidth{0.4pt}
\begin{table}[]
\caption{Optimizers and hyperparameters used for our experiments}\label{table:optimizers}
\centering
 \renewcommand{\arraystretch}{1}
\begin{tabular}{c|c|c|cc}
\hline
\multirow{2}{*}{Model}                                                                       & \multirow{2}{*}{k-Shot} & \multirow{2}{*}{Optimizer} & \multicolumn{2}{c}{Learning Rate} \\
                                                                                             &                         &                            & Feature Extractor   & Classifier  \\ \hline
\multirow{2}{*}{C64F}                                                                        & 1                       & \multirow{11}{*}{Adam}     & 1E-03               & 1E-03       \\
                                                                                             & 5                       &                            & 1E-05               & 1E-02       \\ \cline{1-2} \cline{4-5} 
\multirow{2}{*}{C64F-Dropout}                                                                & 1                       &                            & 1E-04               & 5E-05       \\
                                                                                             & 5                       &                            & 1E-05               & 1E-03       \\ \cline{1-2} \cline{4-5} 
\multirow{2}{*}{ResNetS}                                                                     & 1                       &                            & 1E-04               & 1E-02       \\
                                                                                             & 5                       &                            & 1E-04               & 1E-02       \\ \cline{1-2} \cline{4-5} 
\multirow{5}{*}{\begin{tabular}[c]{@{}c@{}}ResNet10\\ \&\\ ResNet10\\ -Dropout\end{tabular}} & 1                       &                            & 1E-05               & 1E-03       \\
                                                                                             & 2                       &                            & 1E-05               & 1E-03       \\
                                                                                             & 5                       &                            & 1E-05               & 1E-03       \\
                                                                                             & 10                      &                            & 1E-05               & 1E-03       \\
                                                                                             & 20                      &                            & 1E-05               & 1E-03       \\ \hline
\end{tabular}%
\end{table}

\appendices
\section{}
Since the performance of neural networks depends on the choice of optimizers and hyperparameters, it is common to search for proper values from validation examples of a dataset. Thus, in Table \ref{table:optimizers}, we provide the optimizers and the hyperparameters where we used for our experiments. For all experiments, we used the Adam \cite{adam} optimizer.

%


\ifCLASSOPTIONcaptionsoff
  \newpage
\fi



%


\bibliographystyle{IEEEtran}
\bibliography{TNNLS_Bib}

%

\begin{IEEEbiography}
[{\includegraphics[width=1in,height=1.25in,clip,keepaspectratio]{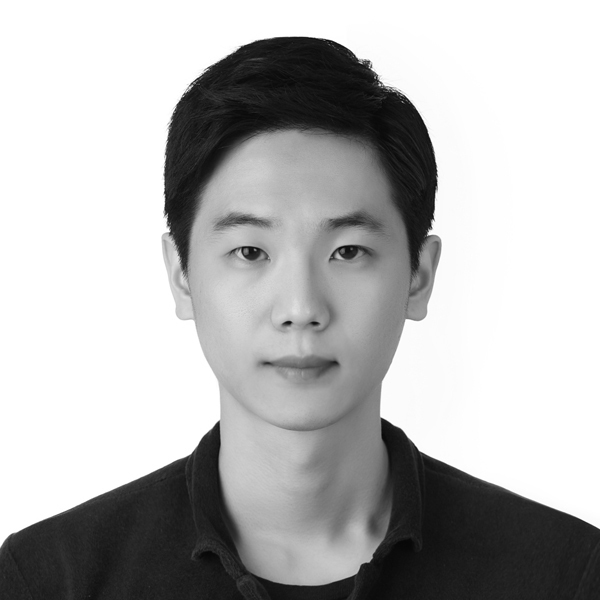}}]{Hong-Gyu Jung} received a B.S. degree in Electronic Engineering and a M.S. degree in Information Telecommunication Engineering from Soongsil University, Seoul,
Korea, in 2012 and 2014, respectively. He is currently a Ph.D. student in the Department of Brain and Cognitive Engineering at Korea University. His current research interests include artificial intelligence and pattern recognition.
\end{IEEEbiography}

\vspace{-2.5mm}

\begin{IEEEbiography}[{\includegraphics[width=1in,height=1.25in,clip,keepaspectratio]{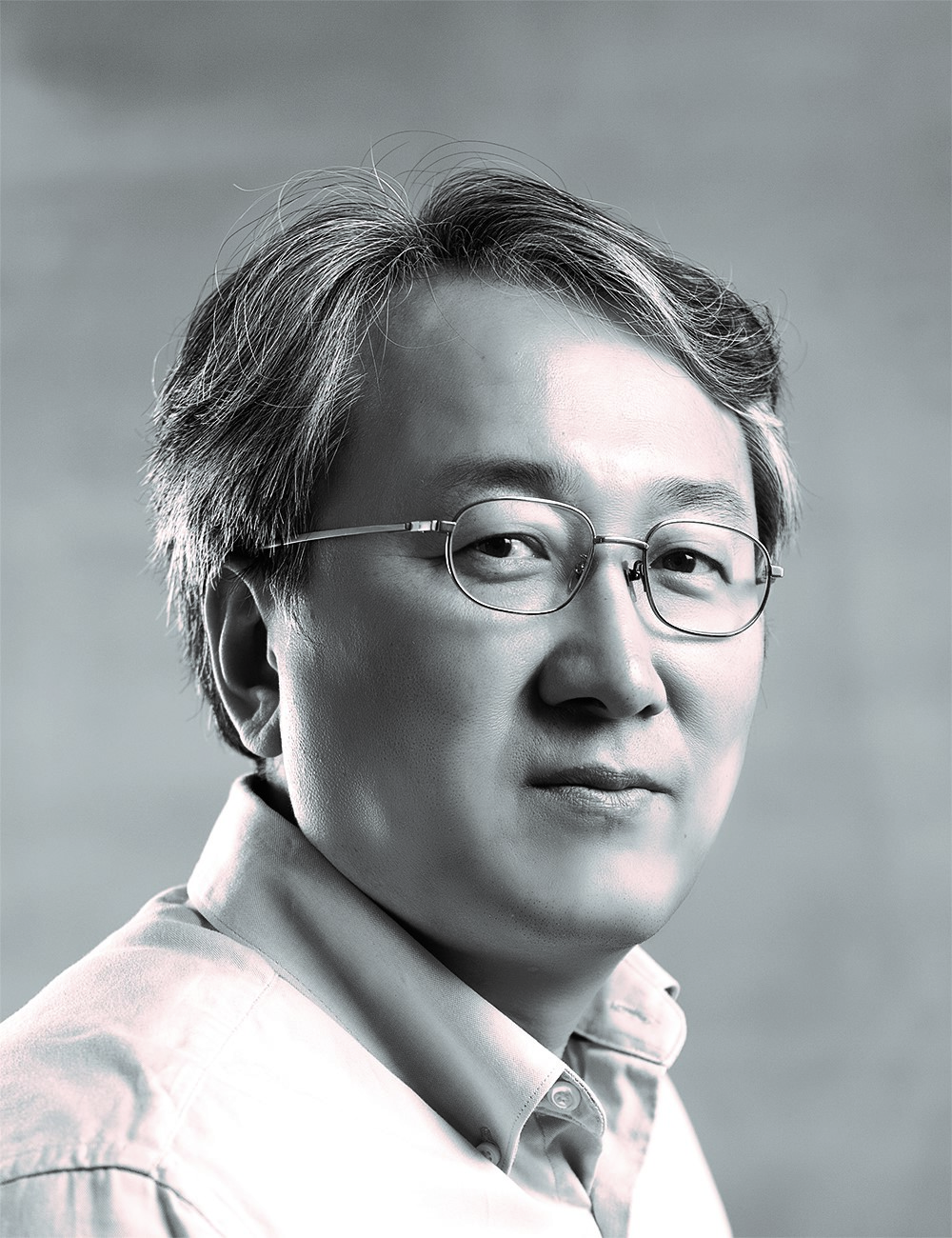}}]{Seong-Whan Lee}
	(S84M89SM96-F10) received a B.S. degree in computer science and statistics from Seoul National University, Korea, in 1984, and M.S. and Ph.D. degrees in computer science from the Korea Advanced Institute of Science and Technology, Korea, in 1986 and 1989, respectively. He is currently the head of the Department of Artificial Intelligence, Korea University. His current research interests include artificial intelligence, pattern recognition, and brain engineering. He is a fellow of the IAPR and the Korea Academy of Science and Technology.
\end{IEEEbiography}




\end{document}